%% file: arxiv.tex
\documentclass[lettersize,journal]{IEEEtran}

\input{preamble}

\myexternaldocument{supplementary}

\begin{document}

\title{\paperTitle}
\author{
    Yang Chen\textsuperscript{*}, 
    Yanbin Wei\textsuperscript{*}, 
    Ke Jin, 
    Yi Kong,
    James T. Kwok, \IEEEmembership{Fellow,~IEEE}, 
    and Yu Zhang, \IEEEmembership{Member,~IEEE}

\thanks{Yang Chen and Yu Zhang are with the Department of Computer Science and Engineering, Southern University of Science and Technology, Shenzhen 518055, China (e-mail: cheny2023@mail.sustech.edu.cn; yu.zhang.ust@gmail.com).}

\thanks{Yanbin Wei and James T. Kwok are with the Department of Computer Science and Engineering, Hong Kong University of Science and Technology (e-mail: yanbin.ust@gmail.com; jamesk@cse.ust.hk).}

\thanks{Ke Jin and Yi Kong are with the School of Information and Control Engineering, China University of Mining and Technology, Xuzhou 221116, China (e-mail: jkpopo97@163.com; kongyicumt@163.com).}

\thanks{\textsuperscript{*} Equal contribution.}

\thanks{Corresponding author: Yu Zhang.}
}

\maketitle

\begin{abstract}
Recent advances in pre-trained vision-language models have demonstrated remarkable zero-shot generalization capabilities. To further enhance these models' adaptability to various downstream tasks, prompt tuning has emerged as a parameter-efficient fine-tuning method. However, despite its efficiency, the generalization ability of prompt remains limited. In contrast, label smoothing (LS) has been widely recognized as an effective regularization technique that prevents models from becoming over-confident and improves their generalization. This inspires us to explore the integration of LS with prompt tuning. However, we have observed that the vanilla LS even weakens the generalization ability of prompt tuning. To address this issue, we propose the \textbf{A}lternating \textbf{T}raining-based \textbf{La}bel \textbf{S}moothing (\textbf{ATLaS}) method, which alternately trains with standard one-hot labels and soft labels generated by LS to supervise the prompt tuning. 
Moreover, we introduce two types of efficient offline soft labels, including Class-wise Soft Labels (CSL) and Instance-wise Soft Labels (ISL), to provide inter-class or instance-class relationships for prompt tuning. 
The theoretical properties of the proposed ATLaS method are analyzed. 
Extensive experiments demonstrate that the proposed ATLaS method, combined with CSL and ISL, consistently enhances the generalization performance of prompt tuning. Moreover, the proposed ATLaS method exhibits high compatibility with prevalent prompt tuning methods, enabling seamless integration into existing methods.

\end{abstract}


\input{1_intro}
\input{2_related}
\input{3_method}
\input{4_experiment}
\input{5_conclusion}

\section*{Acknowledgments}

This work is supported by NSFC grant (under No. 62136005).

\bibliographystyle{IEEEtran}
\bibliography{reference}

\clearpage

\input{6_appendix} 

\end{document}

%% file: preamble.tex
\usepackage{amsmath,amsfonts}
\usepackage{array}

\usepackage[caption=false,font=normalsize,labelfont=sf,textfont=rm]{subfig}
\usepackage{textcomp}
\usepackage{stfloats}
\usepackage{url}
\usepackage{verbatim}
\usepackage{graphicx}
\def\BibTeX{{\rm B\kern-.05em{\sc i\kern-.025em b}\kern-.08em
    T\kern-.1667em\lower.7ex\hbox{E}\kern-.125emX}}
\usepackage{balance}

\usepackage{amssymb}	
\usepackage{algorithm}
\usepackage{algorithmicx}
\usepackage{algpseudocode}
\usepackage{booktabs}
\usepackage{times}
\usepackage{microtype}
\usepackage{epsfig}
\usepackage[inkscapelatex=false]{svg}
\usepackage{float}
\usepackage{placeins}
\usepackage{color, colortbl}
\usepackage{bbding}
\usepackage{pifont}
\usepackage{enumitem}
\usepackage{tabularx}
\usepackage{xstring}
\usepackage{multirow}
\usepackage{xspace}
\usepackage[hang,flushmargin]{footmisc}
\usepackage{mathtools}
\usepackage{xr-hyper}

\makeatletter
\newcommand*{\addFileDependency}[1]{
  \typeout{(#1)}
  \@addtofilelist{#1}
  \IfFileExists{#1}{}{\typeout{No file #1.}}
}

\makeatother
\newcommand*{\myexternaldocument}[1]{
    \externaldocument{#1}
    \addFileDependency{#1.tex}
    \addFileDependency{#1.aux}
}

\newtheorem{assumption}{Assumption}
\newtheorem{theorem}{Theorem}

\newtheorem{lemma}{Lemma}
\newcolumntype{L}[1]{>{\raggedright\arraybackslash}p{#1}}

\definecolor{Gray}{gray}{0.94}

\usepackage{hyperref}
\hypersetup{
colorlinks=true,
linkcolor=blue,
anchorcolor=blue,
citecolor=blue}
\usepackage{cleveref}
\crefname{section}{Section}{Sections}
\crefname{figure}{Fig.}{Figs.}
\crefname{table}{Table}{Tables}

\def\paperTitle{Alternating Training-based Label Smoothing Enhances Prompt Generalization}

%% file: 1_intro.tex
\section{Introduction}
\label{sec:intro}
\IEEEPARstart{R}{ecent} advances in visual-language models (VLMs), as exemplified by CLIP \cite{clip}, ALIGN \cite{align}, BLIP \cite{blip}, EVA-CLIP \cite{evaclip}, LLaVA \cite{llava}, and MiniGPT-4 \cite{minigpt}, have significantly narrowed the long-standing gap between visual perception and language understanding.
By leveraging large-scale image-text pairs during pre-training, 
those VLMs exhibit remarkable capabilities on various real-world visual understanding tasks such as classification  \cite{clip,classification}, object detection \cite{detection,detecion2}, segmentation \cite{segmentation,segmentation2}, captioning \cite{captioning}, visual question answering \cite{flamingo, blip, qwen, qwen2}, and visual reasoning \cite{llava}.

In recent years, prompt tuning \cite{coop, TIP1, TIP2, TIP3} has emerged as a parameter-efficient approach to fine-tune VLMs. The introduction of a few learnable prompts 
to a frozen pre-trained 
VLM (such as CLIP) allows it to adapt to various downstream tasks. However, those prompts tend to overfit the task-specific data distributions, raising concerns about their generalization capabilities across diverse tasks.

To address the limited generalization capability of prompts, we consider the label smoothing (LS) \cite{ls, whendoesls, revisitingls}, a widely recognized regularization technique.
Originally introduced to mitigate the over-fitting in classification tasks, LS reduces the over-confidence of a model by distributing a small portion of probability mass across incorrect labels. LS has been shown to improve the generalization by encouraging models to produce softer and more calibrated predictions \cite{ols, zipf, labelnoise}.
Given its effectiveness in reducing over-fitting and enhancing generalization, this motivates us to explore the integration of LS with prompt tuning.

\begin{figure}[tbp]
    \centering
    \includegraphics[width=0.8 \linewidth]{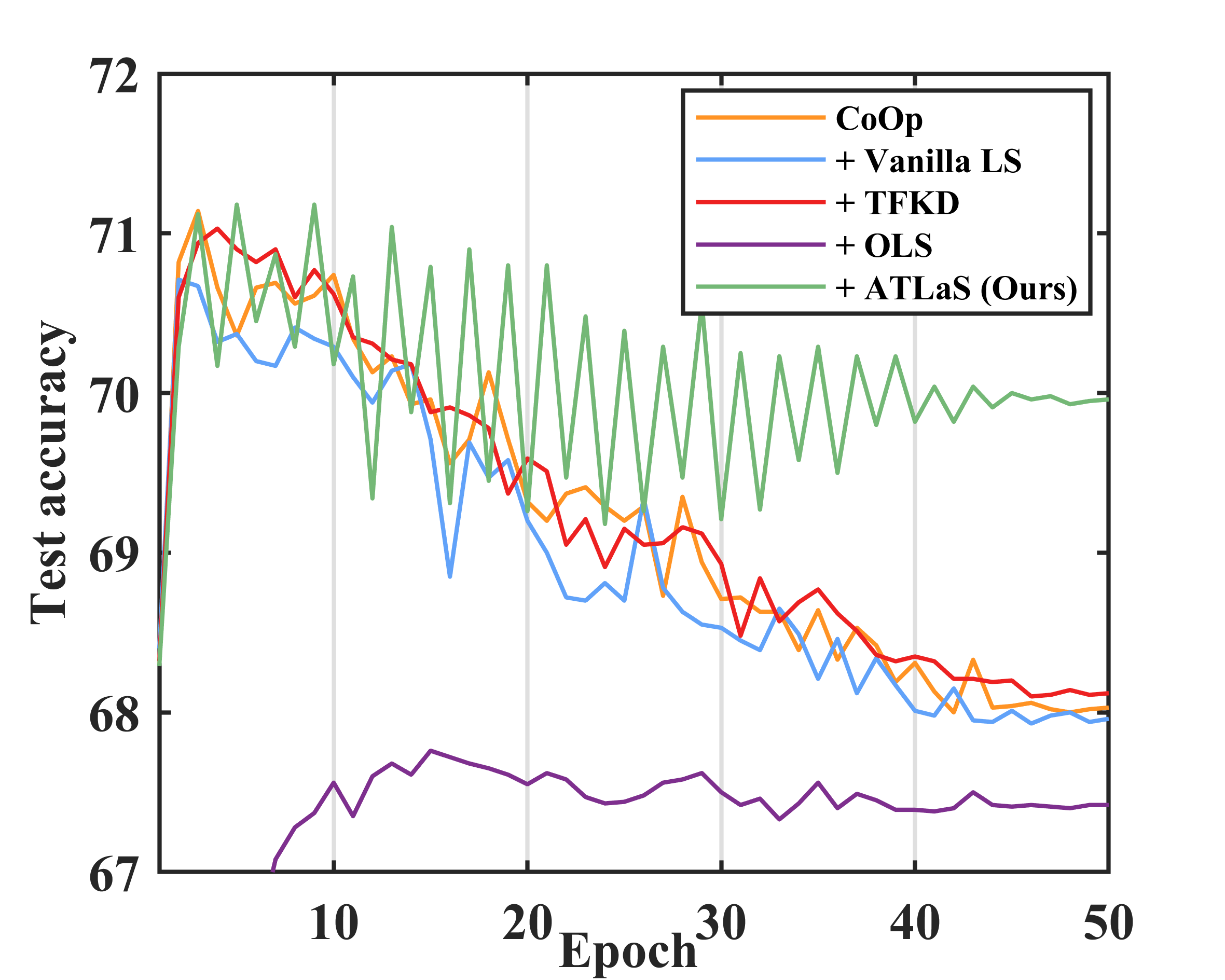}
    \caption{Comparison of test accuracies among CoOp \cite{coop}, CoOp with vanilla label smoothing (LS) \cite{ls}, TFKD \cite{revisitingls}, OLS \cite{ols}, and alternating training-based label smoothing (ATLaS), where prompts are trained on the base classes of ImageNet and subsequently evaluated on new classes.}
    \label{fig:introduction}
\end{figure}

However, our preliminary investigation reveals that employing the LS strategy for prompt tuning fails to enhance the generalization capability of prompts, which aligns with an observation in \cite{dowe} that ``label smoothing doesn't improve fine-tuning of a pre-trained model''. 
Specifically, we conduct preliminary experiments using a representative prompt tuning method, CoOp \cite{coop}, as a baseline, where experimental settings are described in \cref{base2new} of the base-to-new generalization part.
As illustrated in \cref{fig:introduction}, CoOp with vanilla LS demonstrates slightly inferior testing performance compared to the CoOp method throughout the prompt tuning process.
The variants of LS do not improve the testing performance (i.e., TFKD \cite{revisitingls}), but rather impair it (i.e., OLS \cite{ols}).
Naturally, we raise a question: ``Can the LS be revised to help improve the generalization capacity of prompt tuning''.

To answer this question, we propose an \textbf{A}lternating \textbf{T}raining-based \textbf{La}bel \textbf{S}moothing (\textbf{ATLaS}) method, a novel label smoothing technique, for prompt tuning. Specifically, ATLaS alternates between using one-hot labels and soft labels as the supervision information to guide prompt tuning, thereby enhancing the generalization capability of learnable prompts. 
\cref{fig:introduction} shows that CoOp with ATLaS successfully enhances the prompt generalization.
For the generation of soft labels in VLMs, we propose two efficient methods. The first method produces Class-wise Soft Labels (CSL) to derive inter-class similarities from textual prompts, and the second method generates Instance-wise Soft Labels (ISL) to capture instance-class associations from the interaction between textual prompts and instances. Those two types of soft labels could effectively guide the learning of prompts by providing rich supervision signals from different perspectives. 
Theoretically, we analyze ATLaS from the optimization perspective, demonstrating its advantages over the vanilla LS. Extensive experiments demonstrate that the proposed ATLaS method effectively enhances the generalization of prompts in several settings (i.e., cross-dataset transfer, domain generalization, base-to-new, and few-shot classification).

The main contributions of this work are four-fold.
\begin{itemize}
\item To the best of our knowledge, this is the first attempt to successfully integrate LS into prompt tuning to improve the generalization capacity. To address the ineffectiveness of vanilla LS in prompt tuning, we propose the ATLaS method to guide prompt tuning by alternating between one-hot labels and soft labels.
\item We introduce two types of soft labels to guide prompt tuning, including class-wise and instance-wise soft labels, both of which can be efficiently generated before the training process. 

\item The convergence properties of the proposed ATLaS method are analyzed.

\item Extensive experiments demonstrate the effectiveness of the proposed ATLaS method.
\end{itemize}

%% file: 2_related.tex
\section{Related Work}
\label{sec:related}
\subsection{CLIP-based VLMs} VLMs \cite{clip,blip,flamingo,llava} have significantly enhanced cognitive capabilities by integrating visual and textual modalities, excelling at real-world vision understanding tasks such as image classification. Among those models, CLIP \cite{clip} stands out as a foundational pioneer, establishing the standard practice of aligning visual and textual embeddings within a unified representation space.
Given CLIP's foundational role in inspiring a broad family of subsequent models, including ALIGN \cite{align}, EVA-CLIP \cite{evaclip}, OpenCLIP \cite{openclip}, and LongCLIP \cite{longclip}, we use CLIP as the foundational VLM to explore the novel prompt tuning paradigm with label smoothing.

\subsection{Prompt Tuning for VLMs}
As a lightweight model adaptation method, prompt tuning has attracted widespread attention in VLMs \cite{prograd, kgcoop, promptsrc, mopd}. Unlike traditional fine-tuning methods that update all parameters of the pre-trained model, prompt tuning optimizes only the prompt representations, thereby reducing the computational costs. In particular, CoOp \cite{coop} achieves efficient adaptation by optimizing a set of learnable prompt vectors, which replace the hand-crafted textual prompts (e.g., ``a photo of panda'') at the language branch of CLIP. However, despite its simplicity and effectiveness, CoOp tends to overfit base classes, exhibiting limited generalization ability. To address this limitation, CoCoOp \cite{cocoop} learns a lightweight meta-net that generates input-conditional vectors for each input image. ProGrad \cite{prograd} regulates the gradient direction to mitigate conflicts between hand-crafted prompts and learnable prompts. MaPLe \cite{maple} introduces a multi-modal prompt learning method that simultaneously learns hierarchical prompts for both vision and language branches to improve the alignment between the visual and textual representations. To address the base-new trade-off issue, DePT \cite{dept} decouples base-specific knowledge and task-shared knowledge in feature channels.

\subsection{Label Smoothing}
Empirical evidence proves that applying LS strategies during training improves the generalization of the model \cite{ls}.
LS is a regularization strategy that replaces one-hot labels with soft labels, preventing the model from becoming over-confident in its predictions. 
Müller et al. \cite{whendoesls} demonstrate that LS encourages samples from the same class to group in tight clusters, which results in a loss of information in the logits regarding the similarities of different classes, consequently impairing knowledge distillation.
To generate more reliable soft labels, Online label smoothing (OLS) \cite{ols} derives class-wise soft labels online by retaining historical predictions. 
Yuan et al. \cite{revisitingls} explore the intrinsic relationship between LS and knowledge distillation, and propose a teacher-free knowledge distillation (TFKD) framework, where a student model engages in self-learning or draws insights from manually designed regularization distribution.
Different from existing LS methods whose optimization paradigm is to jointly optimize a designed loss of label smoothing along with the standard cross-entropy loss, the proposed ATLaS method alternately optimizes the model with the loss of label smoothing and the standard cross-entropy loss, achieving better performance.


%% file: 3_method.tex
\section{Methodology}
\label{sec:method}

In this section, we first briefly review CLIP, CoOp, and label smoothing by introducing their formulations, and then present the proposed ATLaS method.

\subsection{Preliminaries}
\noindent{\bf Contrastive Language-Image Pre-training (CLIP)} \cite{clip} is a VLM pre-trained on about 400 million image-text pairs, resulting in remarkable zero-shot image recognition capability. CLIP enables zero-shot image classification by measuring the similarity between visual and textual embeddings. Specifically, CLIP consists of a visual encoder and a text encoder. Let $\mathbf{I}$ be the visual embedding extracted by the visual encoder in CLIP for an image $\mathbf{x}$, and $\{ \mathbf{t}_c \} _{c=1}^{C}$ be a set of textual embeddings generated by the text encoder for textual prompts, where
$C$ is the number of classes. 
Those prompts typically follow the format like ``a photo of a [CLS]'', where ``[CLS]'' denotes a class token which is subsequently replaced by specific class names such as ``panda'', ``dog'', and ``bird''. 
Given those embeddings, the prediction probability for image classification can be computed as
\begin{equation}
p(y|\mathbf{x})=\frac{\exp\mathrm{(}f_y\left( \mathbf{x} \right) /\tau )}{\sum\nolimits_{c=1}^C{\exp\mathrm{(}f_c\left( \mathbf{x} \right) /\tau )}},
\label{eq_p_hard}
\end{equation}
where $\tau$ denotes the temperature, $f_c\left( \mathbf{x} \right) =\cos \left( \mathbf{t}_c,\mathbf{I} \right)$, and
$\cos(\cdot,\cdot)$ denotes the cosine similarity.

\noindent {\bf Context Optimization (CoOp)} \cite{coop} is a prompt tuning method designed to replace textual prompts with learnable prompts, enabling VLMs to adapt to downstream tasks through parameter-efficient fine-tuning with a few labeled samples. Specifically, CoOp introduces learnable prompt vectors $\mathbf{v} = \{ \mathbf{v}_1,\mathbf{v}_2,\dots ,\mathbf{v}_M\}$ to replace the textual prompt (e.g., ``a photo of a''), where each $\mathbf{v}_i$ has the same dimension as the word embeddings. The learnable prompt is subsequently constructed by concatenating $\mathbf{v}$ with the class token. Formally, the learnable prompts are formed as $\{ \mathbf{v}_1,\mathbf{v}_2,\dots ,\mathbf{v}_M,[ \mathrm{CLS} ] \} $. Let $\{\mathbf{t}_{c}^{\mathrm{s}}\}_{c=1}^{C}$ be a set of textual embeddings generated by the text encoder for the learnable prompts. The learnable prompt can be optimized by minimizing the cross-entropy loss as
\begin{equation}
\ell \left( \mathbf{y},f\left( \mathbf{v};\mathbf{x} \right) \right) =-\sum_{i=1}^C{y_i\log \frac{\exp\mathrm{(}f_{i}^{\mathrm{s}}\left( \mathbf{v};\mathbf{x} \right) /\tau )}{\sum\nolimits_{j=1}^C{\exp\mathrm{(}f_{j}^{\mathrm{s}}\left( \mathbf{v};\mathbf{x} \right) /\tau )}}},
\label{eq_ce}
\end{equation}
where $f_{c}^{\mathrm{s}}( \mathbf{v};\mathbf{x} ) =\cos ( \mathbf{t}_{c}^{\mathrm{s}},\mathbf{I} ) $, and $y_i$ equals 1 when class $i$ is the ground-truth label for $\mathbf{x}$ and 0 otherwise. Note that the image and text encoders remain frozen during training.




\noindent {\bf Label Smoothing (LS)} \cite{ls} is a regularization strategy that uses soft labels, generated by the weighted averaging of a uniform distribution and hard labels, as a substitute for hard labels during the training process. This technique is designed to mitigate the model over-confidence, thereby enhancing the generalization capabilities of the model. Specifically, LS modifies the one-hot label with the soft label $\mathbf{y}^{\mathrm{LS}}$ as
\begin{equation}
\mathbf{y}^{\mathrm{LS}}=\left( 1-\theta \right) \mathbf{y}+\theta \hat{\mathbf{y}},
\label{eq_ls}
\end{equation}
where $\theta\in[0,1] $ is a smoothing parameter, 
$\mathbf{y}$ denotes the one-hot label, and 
\begin{equation} \label{eq:yhat}
\hat{\mathbf{y}} \sim
\mathbb{P} _{\hat{\mathbf{y}}}
\end{equation} 
is drawn from a distribution $\mathbb{P} _{\hat{\mathbf{y}}}$. Following existing works \cite{ls,whendoesls,revisitingls}, we consider a uniform distribution across all $C$ classes for $\hat{\mathbf{y}}$, i.e., $\hat{\mathbf{y}}={\frac{1}{C}}\mathbf{1}_C$, where $\mathbf{1}_C\in \mathbb{R} ^C$ is the vector of all ones.

\begin{figure*}[t]
    \centering
    \includegraphics[width=\linewidth]{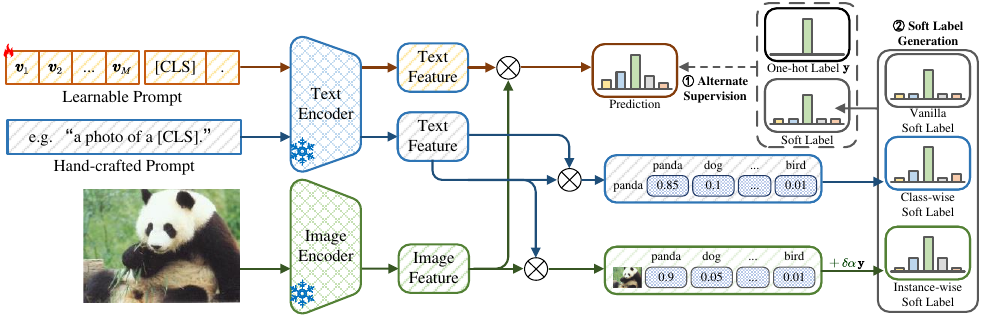}
    \caption{Overview of the proposed ATLaS method. \ding{172} ATLaS supervises the prompt tuning process by alternating between one-hot labels and soft labels, where soft label supervision follows every $K-1$ epochs of one-hot label training. \ding{173} While the basic ATLaS employs vanilla label smoothing for soft label generation, we further exploit the multimodal properties of CLIP to offer two additional options: class-wise soft labels and instance-wise soft labels.}
    \label{fig:overview}
\end{figure*}

\subsection{Alternating Training-based Label Smoothing}

The prompts learned in CoOp exhibit limited generalization capabilities \cite{cocoop}. As the LS strategy is specifically proposed to enhance model generalization, this naturally motivates the integration of LS into the prompt tuning framework to enhance the generalization of the resultant prompts.

However, as discussed in the introduction (i.e., \cref{fig:introduction}), directly using vanilla LS fails to improve the generalization performance and even impairs it.
One possible reason is that the pre-trained CLIP model is trained with one-hot labels but not soft labels, making LS integrated with prompt tuning inconsistent with such a training process and thereby weakening the generalization capacities.

To alleviate the inconsistency of labels used in the pre-training and fine-tuning phases as discussed above, the proposed ATLaS method illustrated in \cref{fig:overview} utilizes both the one-hot labels and soft labels.
Specifically, ATLaS trains with the one-hot labels and soft labels alternately in $K$ successive training epochs, where the construction of soft labels will be introduced in Section \ref{sec:sof_labels}. That is, in $K$ successive training epochs, the ATLaS method learns the prompts using one-hot labels $\mathbf{y}$ in the first \( (K-1) \) epochs and then trains with 
the soft labels \( \mathbf{y}^{\mathrm{LS}} \) in the $K$-th epoch. This alternating training strategy continues throughout the training process. 

\subsection{Analysis}
\label{subsec:analysis}

To understand ATLaS from the optimization perspective, in this section, we provide convergence analysis for ATLaS. 

Let $f(\mathbf{v}; \mathbf{x})$ be the classification model parameterized by the learnable prompt vectors $\mathbf{v}$ to output classification probabilities. 
The expected loss is denoted by $F(\mathbf{v}) = \mathbb{E}_{(\mathbf{x},\mathbf{y})}[\ell(\mathbf{y},f(\mathbf{v};\mathbf{x}))]$, where $\ell(\cdot,\cdot)$ denotes the cross-entropy loss, and $\mathbf{y}$ denotes the one-hot label of $\mathbf{x}$.

We present the following assumptions about smoothness and bounded variance of stochastic gradients, which are commonly used in the non-convex optimization literature \cite{ghadimi2013stochastic, reddi2016stochastic, yan2018unified}.

\begin{assumption}
\label{assump:1}
    (Smoothness). 
    $F(\mathbf{v})$ is $\beta$-smooth in $\mathbf{v}$, i.e., $\|\nabla_{\mathbf{v}} F(\mathbf{v})-\nabla_{\mathbf{v}} F(\mathbf{u})\| \leq \beta\|\mathbf{v}-\mathbf{u}\|$.
\end{assumption}

\begin{assumption}
\label{assump:2}
    (Bounded variance of stochastic gradients). 
    The variance
    of the stochastic gradient is bounded, i.e., there exists a constant $\sigma^
    2 > 0$, such that
    \(
\mathbb{E}_{\left(\mathbf{x},\mathbf{y}\right)}\left\|\nabla_{\mathbf{v}} F(\mathbf{v})-\nabla_{\mathbf{v}}\ell\left(\mathbf{y},f\left(\mathbf{v};\mathbf{x}\right)\right)\right\|^2 \leq \sigma^2
    \).
\end{assumption}


Based on 
Assumption \ref{assump:2} (which is on
stochastic gradient with one-hot label $\mathbf{y}$),
we have the following lemma.\footnote{All the proofs are put in the appendix.} 
\begin{lemma}
\label{lemma:1}
    \(    \mathbb{E}_{(\mathbf{x}, \mathbf{\hat{y}})} \left\| \nabla_{\mathbf{v}} F(\mathbf{v}) - \nabla_{\mathbf{v}} \ell(\mathbf{\hat{y}}, f(\mathbf{v}; \mathbf{x})) \right\|^2 \leq \hat{\sigma}^2 = \kappa \sigma^2
    \)
where \(\kappa > 0\) is a constant, and \(\sigma^2\) is the variance described in Assumption \ref{assump:2}.
\end{lemma}

With the above assumptions and Lemma \ref{lemma:1}, we can establish the convergence bound of the proposed ATLaS method in Theorem \ref{theorem_1}.

\begin{theorem}
\label{theorem_1}
With Assumptions \ref{assump:1} and \ref{assump:2}, 
learning rate 
$\eta = \frac{1}{\beta}$ 
and 
label smoothing parameter
$\theta$, ATLaS satisfies
\[ \frac{1}{T}\sum_{t=0}^{T-1}
\mathbb{E}[\| \nabla_{\mathbf{v_t}} F(\mathbf{v}_t) \|^2] \leq \frac{2F(\mathbf{v}_0)}{\eta T} +\left(1-\frac{\theta}{K}+\frac{\theta\kappa}{K}\right)\sigma^2,
\]
where $T$ is the total number of iterations in the optimization process and $\mathbf{v}_t$ is the prompt vectors at iteration $t$.
\end{theorem}

As articulated in Theorem \ref{theorem_1}, a smaller $\kappa$, which is associated with the quality of $\hat{\mathbf{y}}$ employed in LS, facilitates faster convergence. 
It is noteworthy that the convergence bound for vanilla LS is \(\frac{2F(\mathbf{v}_0)}{\eta T} + (1-\theta+\theta\kappa)\sigma^2\) \cite{xu2020towards},\footnote{We provide a detailed convergence analysis of label smoothing by \cite{xu2020towards} in Appendix \ref{appendix:proof_ls}.} which is unequivocally larger than that for ATLaS when $\kappa > 1$. This suggests that ATLaS can converge better during the optimization of parameters where the selection of $\hat{\mathbf{y}}$ is suboptimal (i.e., characterized by a large $\kappa$).

According to Theorem \ref{theorem_1}, soft labels affect the convergence properties of the proposed ATLaS method via $\kappa$. In the following section, we propose two ways to generate effective soft labels.


\subsection{Generation of Offline Soft Labels}
\label{sec:sof_labels}

Vanilla LS uses a uniform distribution to generate soft labels, treating all non-target categories equally without considering inter-class or instance-class relationships. Such naive soft labels will potentially limit model performance by disregarding inherent structural relationships within the data. Moreover, the acquisition of prior knowledge about inter-class or instance-class relationships is a significant challenge, often requiring domain expertise or extensive data analysis. To generate better soft labels to supervise prompt tuning, we exploit the multimodal capabilities of CLIP to introduce two novel types of offline soft labels: \textbf{Class-wise Soft Labels (CSL)} and \textbf{Instance-wise Soft Labels (ISL)}, which can be generated before the prompt tuning stage.

\textbf{CSL} generates soft labels that encapsulate prior knowledge about inter-class similarities. We derive CSLs from the textual modality through the interaction of textual prompts. Specifically, given a set of textual embeddings $\{ \mathbf{t}_c \} _{c=1}^{C}$ of labels, a similarity matrix $\mathbf{A} \in \mathbb{R}^{C \times C}$ to measure the inter-class similarities is computed before the training phase as 
\begin{equation}
A_{ij}=\frac{\exp \left( \mathrm{Sim}\left( \mathbf{t}_i,\mathbf{t}_j \right) /\tau _c \right)}{\sum\nolimits_{j=1}^C{\exp \left( \mathrm{Sim}\left( \mathbf{t}_i,\mathbf{t}_j \right) /\tau _c \right)}}
\label{eq_a_ij}
\end{equation}
where $A_{ij}$, the $(i,j)$-th entry in $\mathbf{A}$, denotes the class-level similarity between the $i$-th and $j$-th classes, $\mathrm{Sim}( \cdot ,\cdot ) $ denotes the similarity function, and $\tau_c$ is the temperature parameter. Here we use the cosine similarity for $\mathrm{Sim}( \cdot, \cdot ) $. Then the soft label for the $c$-th class in the CSL is defined as the $c$-th row of $\mathbf{A}$, i.e., $\mathbf{y}_{c}^{\mathrm{CSL}}=\mathbf{A}_c$.
Note that each diagonal entry in $\mathbf{A}$ is the largest among the corresponding column. Hence, the $c$-th entry in $\mathbf{y}_{c}^{\mathrm{CSL}}$ is the largest, making CSL a reasonable method of generating soft labels.

Due to inherent factors such as pose, deformation, and lighting conditions, images within the same category can exhibit substantial intra-class variance \cite{diversity1,diversity2}. Such intra-class variance prevents CSL from adequately capturing visual differences.
Moreover, the utilization of knowledge in a single modality to supervise prompt tuning could potentially disrupt the vision-language alignment learned by CLIP. To address those concerns in CSL, we propose the \textbf{ISL}, which utilizes both the textual and visual modalities through the interaction between textual prompts and individual instances to capture instance-class associations. 
Specifically, we can use the prediction probabilities of CLIP as soft labels, which are instance-dependent, and learn the instance-class relations. However, this approach can not guarantee that the value for the ground-truth class in the soft label be the largest. Therefore, the soft label in ISL is rectified to be a combination of the one-hot label $\mathbf{y}$ and the prediction probabilities as
\begin{equation}
\mathbf{y}^{\mathrm{ISL}}=\frac{\mathbf{p}(\cdot |\mathbf{x})+\delta \alpha \mathbf{y}}{1+\delta \alpha} \label{eq_ISL}
\end{equation}
where $\alpha$ is the rectification coefficient, $\mathbf{p}(\cdot |\mathbf{x})=[p(y_1|\mathbf{x}), p(y_1|\mathbf{x}),\dots, p( y_C|\mathbf{x})]^\mathrm{T}$ denotes the prediction probabilities by CLIP, $\mathbb{I}(\cdot)$ is the indicator function, and $\delta =\mathbb{I} ( \mathrm{arg}\max \mathbf{p}( \cdot |\mathbf{x} ) \ne \mathrm{the} \;\mathrm{ground}\mbox{-} \mathrm{truth} \;\mathrm{class} )$. In other words, the rectification process defined in Eq.~(\ref{eq_ISL}) is exclusively implemented when the class corresponding to the maximum value of the soft label diverges from the ground-truth class.

Notably, since both CSL and ISL can be generated offline before the fine-tuning process, they introduce negligible computational cost during the training process. Moreover, the generation of soft labels is not restricted to CLIP and it is easy to extend to advanced VLMs.
Additionally, the proposed ATLaS method with CSL and ISL can be integrated with existing prompt tuning methods without introducing extra learnable parameters.

In summary, the whole algorithm for the proposed ATLaS method is shown in Algorithm \ref{alg:ATLaS}.

\begin{algorithm}
\caption{The ATLaS method}
    \begin{algorithmic}[1]
    \State Initialize $\mathbf{v}_0$, $\theta$, and $\eta$.
    \For {epoch $e = 0, 1, \dots, E - 1$}
    \State $\xi:=\begin{cases}
	0,\ \mathrm{if} \; e+1\equiv 0\left( \mathrm{mod}\;K \right)\\
	1, \mathrm{ otherwise}\\
        \end{cases}$
        \For {each training iteration $t$}
            \State sample a mini-batch $(\mathbf{x}_t, \mathbf{y}_t)$
            \If{using CSL}
                \For {each label $\mathbf{y}=[y_1, y_2,...,y_C]^\mathrm{T}$ in $\mathbf{y}_t$}
                    \State find the index $i$ where $y_i = 1$
                    \State $\mathbf{y}^{\mathrm{LS}}:=\mathbf{y}_{i}^{\mathrm{CSL}}$
                \EndFor
            \ElsIf{using ISL}
                \State $\mathbf{y}_{t}^{\mathrm{LS}}:=\mathbf{y}_{t}^{\mathrm{ISL}}$
            \Else
                \State $\mathbf{y}_{t}^{\mathrm{LS}}:=\left( 1-\theta \right) \mathbf{y}_{t}+\theta \hat{\mathbf{y}}$
            \EndIf
            \State $\mathbf{y}_{t}^{\mathrm{ATLaS}}:=\xi \mathbf{y}_t+( 1-\xi) \mathbf{y}_{t}^{\mathrm{LS}}$
            \State $\mathbf{v}_{t+1}: = \mathbf{v}_{t} - \eta \nabla_\mathbf{v} \ell(\mathbf{y}_{t}^{\mathrm{ATLaS}}, f(\mathbf{v}_{t}; \mathbf{x}_t))$
        \EndFor
    \EndFor
\end{algorithmic}
\label{alg:ATLaS}
\end{algorithm}

\subsection{Analysis of the Soft Labels Generated by CSL and ISL}
\begin{figure}[!tbph]
  \centering
  \includegraphics[width=0.8\linewidth]{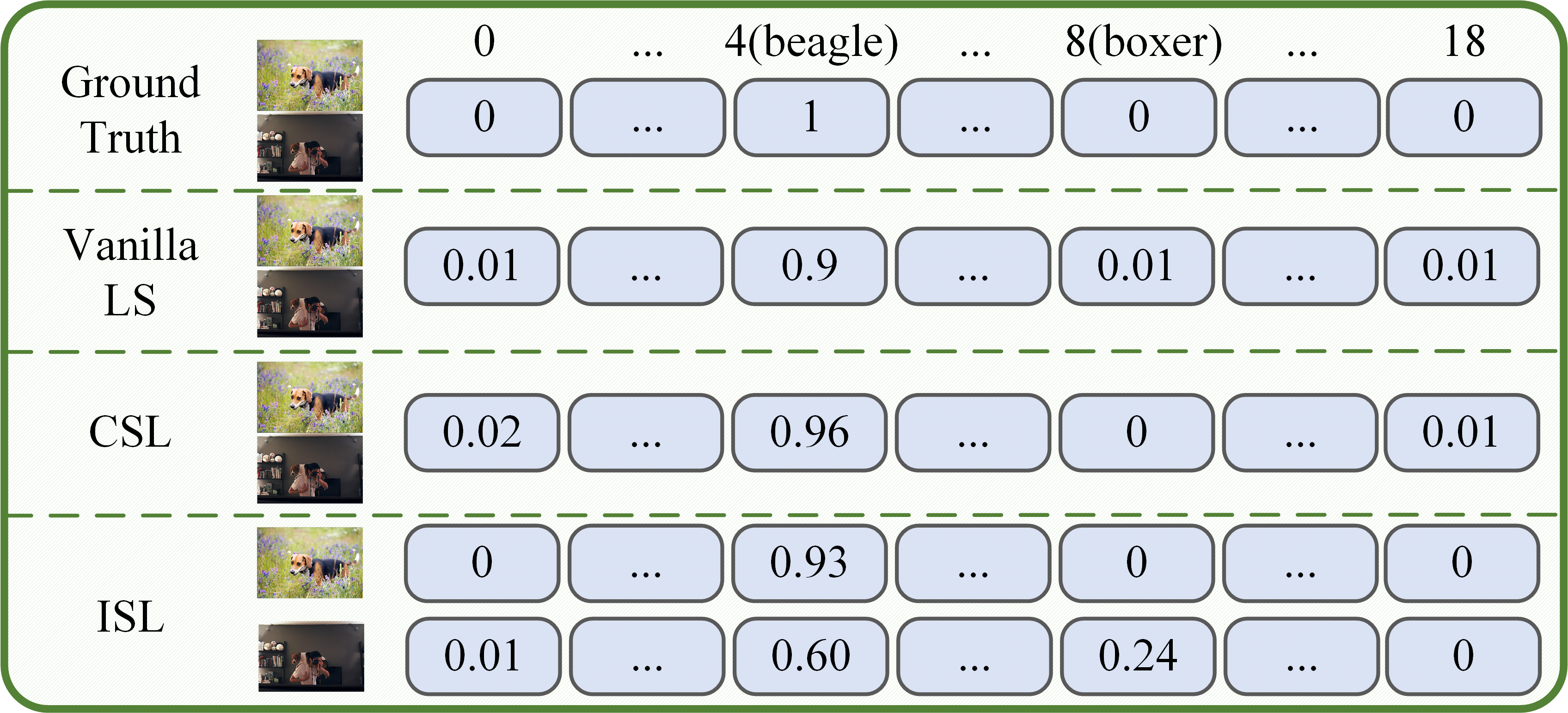}
   \caption{Case study of the soft labels generated by ISL and CSL.}
   \label{soft_label}
\end{figure}

Here we provide a case study with detailed explanations in \cref{soft_label}, showing the soft labels generated by CSL and ISL for two same-class images. 
Firstly, CSL enhances vanilla LS by incorporating inter-class similarity knowledge. While vanilla LS assigns equal value to non-target classes, CSL generates more informed soft labels (e.g., 0.96 confidence for the target class, with varying confidence for non-target classes based on their similarities).
Secondly, ISL generates instance-specific soft labels to capture intra-class variance (e.g., 0.93 and 0.60 confidence for the target class of different images). 
For an image containing both a human and a beagle, ISL effectively captures this contextual information and assigns 0.24 confidence to the `boxer' class, showing its ability to generate more semantically rich soft labels.

%% file: 4_experiment.tex
\section{Experiments}
\label{sec:experiment}

In this section, we empirically evaluate the proposed ATLaS method. 

\subsection{Experimental Setup}

\paragraph{Datasets}
Following CoOp \cite{coop} and CoCoOp\cite{cocoop}, experiments under four settings, including cross-dataset generalization, base-to-new generalization, and few-shot classification, are conducted on 11 image classification datasets. Those datasets include ImageNet \cite{imagenet} and Caltech101 \cite{caltech101} for generic object classification, OxfordPets \cite{oxfordpets}, StanfordCars \cite{stanfordcars}, Flowers102 \cite{oxfordflowers}, Food101 \cite{food}, and FGVCAircraft \cite{FGVCAircraft} for fine-grained visual categorization, SUN397 \cite{sun397} for scene recognition, DTD \cite{dtd} for texture classification, EuroSAT \cite{eurosat} for satellite image classification, and UCF101 \cite{ucf101} for action recognition. Besides, following \cite{cocoop}, experiments under the domain generalization setting are conducted on the ImageNet dataset and its variants, including ImageNetV2 \cite{imagenet_v2}, ImageNet-Sketch \cite{imagenet_sketch}, ImageNet-A \cite{imagenet_a}, and ImageNet-R \cite{imagenet_r}.

\paragraph{Baselines}
We evaluate the effectiveness of the proposed ATLaS method by incorporating it into a broad spectrum of baseline methods, including both textual prompt tuning methods (e.g., CoOp \cite{coop}, Conditional Context Optimization (CoCoOp) \cite{cocoop}, and Decoupled Prompt Tuning (DePT) \cite{dept}) and multi-modal prompt tuning methods (e.g., Independent Vision-Language Prompting (IVLP) \cite{ivlp} and Multi-modal Prompt Learning (MaPLe) \cite{maple}).

\paragraph{Implementation Details}
Our implementation is based on CoOp \cite{coop}. For all baseline methods, we maintain the same experimental settings (e.g., training epochs, training schedules, and data augmentation settings) as specified in their original implementations. All experimental results are averaged over three seeds. All experiments are conducted with a ViT-B/16-based CLIP model. The length of prompt vectors is fixed to 4, and they are initialized with the textual template ``a photo of a''. For the ATLaS method, $K$, $\theta$ in Eq.~(\ref{eq_ls}), $\tau_c$ in Eq.~(\ref{eq_a_ij}), and $\alpha$ in Eq.~(\ref{eq_ISL}) are set to 2, 0.1, 0.05, and 0.1, respectively, by default. More details are provided in Appendix \ref{appendix:imp_detail}.

\subsection{Experimental Results}
To empirically validate the effectiveness of our proposed ATLaS method, we conduct a comprehensive set of experiments across four challenging generalization settings, including cross-dataset generalization, domain generalization, base-to-new generalization, and few-shot classification. We demonstrate that ATLaS and its variants (i.e., ATLaS-CSL that uses the CSL to generate soft labels and ATLaS-ISL that uses ISL to generate soft labels) can be seamlessly integrated into a wide range of prompt tuning methods, including CoOp, CoCoOp, DePT, IVLP, and MaPLe, consistently enhancing their generalization capabilities. The experimental settings for all baselines are identical to those in the original implementations.

\subsubsection{Cross-Dataset Generalization}
In the cross-dataset generalization setting that evaluates whether prompts learned from a source domain can generalize to unseen target domains, prompts are trained using 16-shot samples from each of the 1000 classes on a single source domain (i.e., ImageNet) and then evaluated on the other 10 unseen target datasets. Results in \cref{tab:cross} show that integrating ATLaS and its variants consistently improve the cross-dataset generalization performance over baseline methods. Notably, when integrated with CoOp, ATLaS-CSL achieves the highest average accuracy improvement of 2.10\%, showing the powerful benefit of leveraging inter-class textual similarities for domain transfer. The optimal ATLaS variant varies per baseline, highlighting its flexibility in adapting to different prompt tuning architectures.


\begin{table*}[htbp]
  \centering
  \caption{Performance of various methods under the cross-dataset generalization setting. The number in the parentheses denotes the improvement over the baseline method.}
    \resizebox{\linewidth}{!}{
    \setlength{\tabcolsep}{1mm}{
    \begin{tabular}{l|c|ccccccccccc}
    \toprule
          & Source & \multicolumn{11}{c}{Target} \\
\cmidrule{2-13}          & ImageNet & Caltech101 & OxfordPets & StanfordCars & Flowers102 & Food101 & FGVCAircraft & SUN397 & DTD   & EuroSAT & UCF101 & Avg. \\
    \midrule
    CoOp  & 71.70  & 93.40  & 89.54  & 61.90  & 69.23  & 85.66  & 17.70  & 64.05  & 42.00  & 39.78  & 67.80  & 63.11  \\
    \midrule
     + ATLaS & \textbf{71.79} & 92.81  & 90.01  & 64.20  & 69.50  & 85.52  & 18.16  & 64.92  & 41.90  & 46.93  & 68.03  & 64.20 (+1.09) \\
    \rowcolor{Gray}  + ATLaS-CSL & 71.43  & 93.19  & 89.59  & 65.07  & 69.97  & 85.90  & \textbf{21.86} & 65.63  & \textbf{43.44} & \textbf{49.76} & 67.69  & {\textbf{65.21 (+2.10)}} \\
     + ATLaS-ISL & 71.72  & \textbf{93.62} & \textbf{90.19} & \textbf{65.75} & \textbf{70.35} & \textbf{86.26} & 19.53  & \textbf{66.04} & 43.40  & 41.04  & \textbf{68.48} & 64.47 (+1.36) \\
    \midrule
    CoCoOp & 71.02  & 94.24  & 90.43  & 66.06  & 71.70  & 85.92  & 23.10  & 67.65  & 45.04  & 38.73  & 67.99  & 65.09  \\
    \midrule
     + ATLaS & 70.99  & 93.83  & 90.52  & 66.09  & 71.74  & 86.14  & 23.64  & 67.68  & 44.98  & \textbf{43.21} & 68.25  & 65.61 (+0.52) \\
     + ATLaS-CSL & \textbf{71.05} & \textbf{94.48} & \textbf{90.90} & 66.48  & 72.07  & 85.91  & 23.31  & \textbf{68.07} & 45.33  & 39.77  & 68.65  & 65.50 (+0.41) \\
    \rowcolor{Gray}  + ATLaS-ISL & 70.80  & 94.40  & 90.70  & \textbf{66.80} & \textbf{72.60} & \textbf{86.40} & \textbf{24.40} & 67.50  & \textbf{46.30} & 39.60  & \textbf{69.10} & {\textbf{65.78 (+0.69)}} \\
    \midrule
    DePT  & 72.79  & \textbf{93.96}  & 89.97  & 65.05  & 70.44  & \textbf{86.41}  & 21.12  & 66.42  & 43.14  & 46.30  & 67.51  & 65.03  \\
    \midrule
     + ATLaS & \textbf{72.87} & 93.67  & \textbf{90.32}  & 65.50  & 72.03  & 86.32  & 21.87  & \textbf{67.03}  & 45.04  & 45.83  & 68.62  & 65.62 (+0.59) \\
     + ATLaS-CSL & 72.83  & 93.59  & 90.11  & 65.99  & 71.25  & 86.06  & 23.49  & 66.66  & \textbf{46.34}  & 46.32  & 69.05  & 65.89 (+0.85) \\
    \rowcolor{Gray}  + ATLaS-ISL & 72.83  & 93.79  & 90.16  & \textbf{66.00}  & \textbf{72.07}  & 86.25  & \textbf{24.00}  & 66.87  & 45.15  & \textbf{48.86}  & \textbf{69.71}  & {\textbf{66.29 (+1.25)}} \\
    \midrule
    IVLP  & 72.40  & 92.80  & 90.40  & 64.10  & 68.00  & 85.50  & 20.30  & 66.30  & 43.30  & 41.50  & 67.40  & 63.96  \\
    \midrule
    \rowcolor{Gray}  + ATLaS & \textbf{72.60} & 91.20  & \textbf{90.90} & \textbf{65.30} & 69.90  & 85.90  & 23.30  & \textbf{67.40} & \textbf{46.40} & \textbf{44.60} & 68.60  & {\textbf{65.35 (+1.39)}} \\
     + ATLaS-CSL & 72.40  & \textbf{94.10} & 90.30  & 65.10  & \textbf{71.70} & 85.60  & 23.20  & 66.70  & 43.70  & 42.40  & \textbf{68.80} & 65.16 (+1.20) \\
     + ATLaS-ISL & 71.90  & 93.80  & 90.80  & 64.10  & 71.50  & \textbf{86.20} & \textbf{24.80} & 67.20  & \textbf{46.40} & 37.10  & 68.30  & 65.02 (+1.06) \\
    \midrule
    MaPLe & \textbf{72.01} & 93.59  & 89.64  & 63.85  & 70.73  & 85.71  & 23.85  & 66.28  & 45.51  & 47.16  & \textbf{68.70} & 65.50  \\
    \midrule
     + ATLaS & 71.50  & 93.50  & 90.10  & 64.57  & 71.57  & \textbf{86.20} & 23.87  & 67.10  & 45.97  & 50.17  & 67.83  & 66.09 (+0.59) \\
    \rowcolor{Gray}  + ATLaS-CSL & 71.47  & \textbf{93.67} & 90.37  & 64.83  & 71.50  & 86.13  & 24.03  & 67.17  & \textbf{46.30} & \textbf{50.90} & 68.57  & {\textbf{66.35 (+0.85)}} \\
     + ATLaS-ISL & 71.30  & 93.57  & \textbf{90.67} & \textbf{65.70} & \textbf{71.90} & \textbf{86.20} & \textbf{24.07} & \textbf{67.20} & 46.03  & 47.57  & 68.40  & 66.13 (+0.63) \\
    \bottomrule
    \end{tabular}}}
  \label{tab:cross}%
\end{table*}%

\subsubsection{Domain Generalization}
In the domain generalization setting, prompts are evaluated for the ability to generalize to out-of-distribution data. Specifically, prompts are trained using 16-shot samples from each of the 1000 classes on ImageNet, and then evaluated on four different target domains (i.e., ImageNetV2, ImageNet-Sketch, ImageNet-A, and ImageNet-R). Results in \cref{tab:dg} show that the prompts learned by the proposed ATLaS method and its variants are more domain-generalizable. Those results further substantiate the effectiveness of the proposed ATLaS method in learning more transferable prompt representations.

\begin{table}[htbp]
  \caption{Performance of various methods under the domain generalization setting. The number in the parentheses denotes the improvement over the baseline method.}
  \centering
  \resizebox{\linewidth}{!}{
\setlength{\tabcolsep}{2mm}{
    \begin{tabular}{l|c|cccc|c}
    \toprule
          & Source & \multicolumn{5}{c}{Target} \\
\cmidrule{2-7}          & ImageNet & -V2   & -Sketch & -A    & -R    & Avg. \\
    \midrule
    CoOp  & 71.51  & 64.20  & 47.99  & 49.71  & 75.21  & 59.28  \\
    \midrule
     + ATLaS & \textbf{71.79} & 64.81  & 48.76  & 50.66  & \textbf{76.99} & 60.31 (+1.03) \\
    \rowcolor{Gray}  + ATLaS-CSL & 71.43  & 64.27  & \textbf{49.02} & \textbf{51.24} & 76.85  & {\textbf{60.35 (+1.07)}} \\
     + ATLaS-ISL & 71.72  & \textbf{64.85} & 48.70  & 50.50  & 76.21  & 60.07 (+0.79) \\
    \midrule
    CoCoOp & 71.02  & 64.07  & 48.75  & 50.63  & 76.18  & 59.90  \\
    \midrule
     + ATLaS & 70.99  & 64.21  & 49.13  & 51.23  & 76.99  & 60.39 (+0.49) \\
    \rowcolor{Gray}  + ATLaS-CSL & \textbf{71.05} & \textbf{64.38} & \textbf{49.19} & \textbf{51.41} & \textbf{77.08} & {\textbf{60.52 (+0.62)}} \\
     + ATLaS-ISL & 70.80  & 64.10  & 48.90  & 51.30  & 77.00  & 60.33 (+0.43) \\
    \midrule
    DePT  & 72.79  & 63.94  & 45.79  & 46.51  & 73.34  & 57.40  \\
    \midrule
    \rowcolor{Gray}  + ATLaS & \textbf{72.87} & 64.03  & \textbf{46.90} & 47.59  & \textbf{74.48} & {\textbf{58.25 (+0.86)}} \\
     + ATLaS-CSL & 72.83  & \textbf{64.11} & 46.81  & \textbf{47.63} & 74.19  & 58.19 (+0.79) \\
     + ATLaS-ISL & 72.83  & 64.09  & 46.85  & 47.57  & 74.12  & 58.16 (+0.76) \\
    \midrule
    IVLP  & 72.40  & 65.20  & 48.80  & 48.00  & 76.50  & 59.63  \\
    \midrule
     + ATLaS & \textbf{72.60} & 65.20  & 49.20  & 48.70  & \textbf{76.70} & 59.95 (+0.33) \\
     + ATLaS-CSL & 72.40  & 65.10  & 49.20  & \textbf{49.00} & 76.40  & 59.93 (+0.30) \\
    \rowcolor{Gray}  + ATLaS-ISL & 71.90  & \textbf{65.30} & \textbf{49.50} & 48.80  & 76.60  & {\textbf{60.05 (+0.42)}} \\
    \midrule
    MaPLe & \textbf{72.01} & \textbf{64.70} & 48.66  & 49.21  & 76.31  & 59.72  \\
    \midrule
     + ATLaS & 71.50  & 64.53  & 48.87  & 50.17  & 76.97  & 60.14 (+0.42) \\
     + ATLaS-CSL & 71.47  & 64.53  & 48.83  & \textbf{50.23} & 76.70  & 60.07 (+0.35) \\
    \rowcolor{Gray}  + ATLaS-ISL & 71.30  & 64.53  & \textbf{49.30} & 50.20  & \textbf{77.03} & {\textbf{60.27 (+0.55)}} \\
    \bottomrule
    \end{tabular}}}
  \label{tab:dg}%
  \vspace{-10pt}
\end{table}%

\subsubsection{Base-to-New Generalization}
\label{base2new}
In the base-to-new generalization setting, all classes are divided into disjoint base and new groups, and the prompts are trained on base classes and evaluated on base (seen) and new (unseen) classes. \cref{tab:b2n} shows that the proposed ATLaS method consistently enhances the generalization performance of all baselines, specifically achieving better new-class accuracy and harmonic mean accuracy, where detailed results for each dataset are provided in Appendix \ref{appendix:exp_detail}. Those results indicate that ATLaS, ATLaS-CSL, and ATLaS-ISL can be seamlessly integrated with existing prompt tuning methods to mitigate overfitting on base classes and enhance generalization to new concepts.

\begin{table}[t]
  \centering
  \caption{The average performance of five prompt tuning methods with or without ATLaS, ATLaS-CSL, and ATLaS-ISL on the 11 datasets under the base-to-new generalization setting. `tp' denotes textual prompt tuning, and `mmp' denotes multi-modal prompt tuning. ‘H’ denotes the harmonic mean accuracy. The number in the parentheses denotes the improvement over the baseline method.}
    \resizebox{\linewidth}{!}{
    \setlength{\tabcolsep}{2.3mm}{
    \begin{tabular}{c|L{2cm}|ccc}
    \toprule
    \multirow{2}[4]{*}{Prompting} & \multicolumn{1}{c|}{\multirow{2}[4]{*}{Methods}} & \multicolumn{3}{c}{Avg. acc. over 11 datasets (\%)} \\
\cmidrule{3-5}          &       & Base  & New   & H \\
    \midrule
    \multirow{4}[4]{*}{tp} & CoOp  & 82.69  & 63.23  & 71.66  \\
\cmidrule{2-5}          &   + ATLaS & 82.42  & 65.19  & 72.80 ({+1.14}) \\
          &   + ATLaS-CSL & 82.03  & 68.05  & 74.39 ({+2.73}) \\
          & \cellcolor{Gray} + ATLaS-ISL & \cellcolor{Gray}80.25  & \cellcolor{Gray}73.62  & \cellcolor{Gray}\textbf{76.79 ({+5.13})} \\
    \midrule
    \multicolumn{1}{c|}{\multirow{4}[4]{*}{tp}} & CoCoOp  & 80.47  & 71.69  & 75.83  \\
\cmidrule{2-5}          &   + ATLaS & 80.13  & 73.43  & 76.64 ({+0.81}) \\
          &   + ATLaS-CSL & 79.79  & 73.69  & 76.62 ({+0.79}) \\
          & \cellcolor{Gray}  + ATLaS-ISL & \cellcolor{Gray}79.19  & \cellcolor{Gray}74.37  & \cellcolor{Gray}\textbf{76.70 ({+0.88})} \\
    \midrule
    \multicolumn{1}{c|}{\multirow{4}[4]{*}{tp}} & DePT   & 83.56  & 71.92  & 77.30  \\
\cmidrule{2-5}          &   + ATLaS & 83.81  & 73.67  & 78.42 ({+1.12}) \\
          &   + ATLaS-CSL & 83.76  & 73.85  & 78.49 ({+1.19}) \\
          & \cellcolor{Gray}  + ATLaS-ISL & \cellcolor{Gray}83.80  & \cellcolor{Gray}75.05  & \cellcolor{Gray}\textbf{79.18 ({+1.88})} \\
    \midrule
    \multicolumn{1}{c|}{\multirow{4}[4]{*}{mmp}} & IVLP   & 83.69  & 71.10  & 76.88  \\
\cmidrule{2-5}          &   + ATLaS & 84.39  & 71.90  & 77.64 ({+0.76}) \\
          & \cellcolor{Gray}  + ATLaS-CSL & \cellcolor{Gray}84.21  & \cellcolor{Gray}73.41  & \cellcolor{Gray}\textbf{78.44 ({+1.56})} \\
          &   + ATLaS-ISL & 84.26  & 73.06  & 78.26 ({+1.38}) \\
    \midrule
    \multicolumn{1}{c|}{\multirow{4}[4]{*}{mmp}} & MaPLe  & 83.52  & 73.22  & 78.04  \\
\cmidrule{2-5}          &   + ATLaS & 83.46  & 74.21  & 78.56 ({+0.52}) \\
          &   + ATLaS-CSL & 83.19  & 74.69  & 78.71 ({+0.67}) \\
          & \cellcolor{Gray}  + ATLaS-ISL & \cellcolor{Gray}83.09  & \cellcolor{Gray}75.13  & \cellcolor{Gray}\textbf{78.91 ({+0.87})} \\
    \bottomrule
    \end{tabular}}}
  \label{tab:b2n}%
\end{table}%

\subsubsection{Few-shot Classification}
\begin{figure*}[!tbph]
  \centering
\subfloat{
    \includegraphics[width=0.29\linewidth]{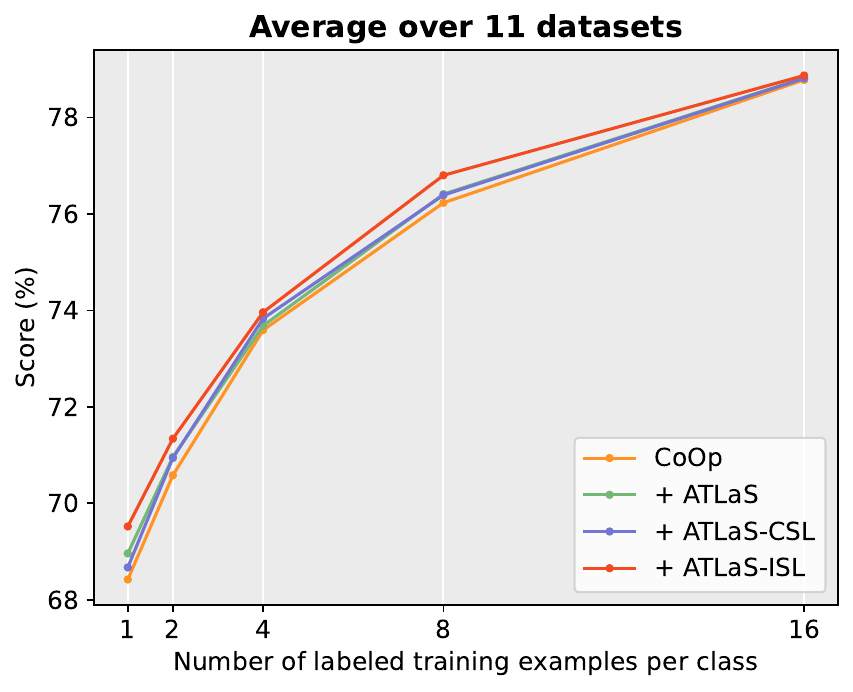}
}
  \hfill
\subfloat{
    \includegraphics[width=0.29\linewidth]{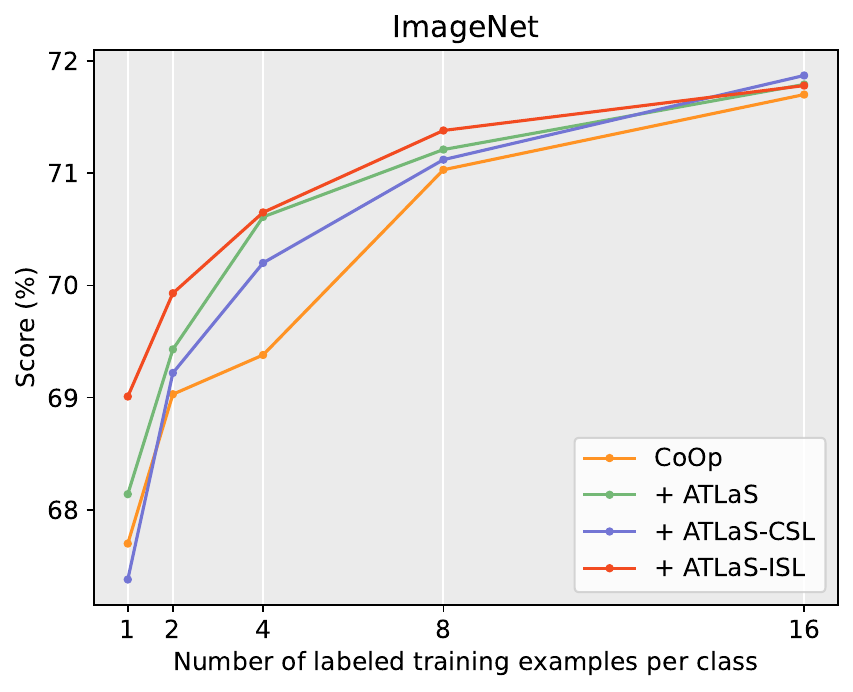}
}
  \hfill
\subfloat{
    \includegraphics[width=0.29\linewidth]{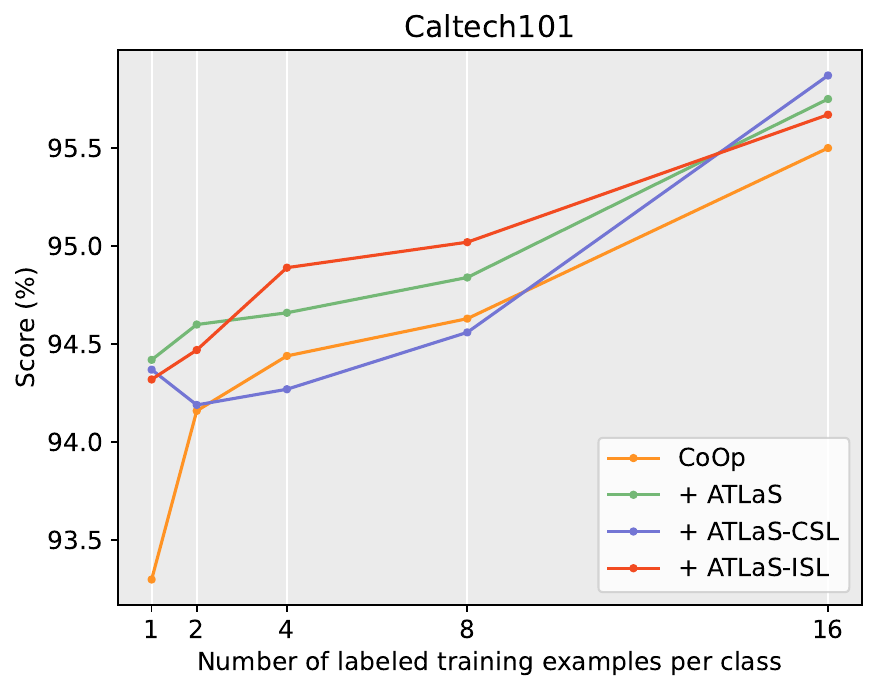}
}
    
  \subfloat{
    \includegraphics[width=0.29\linewidth]{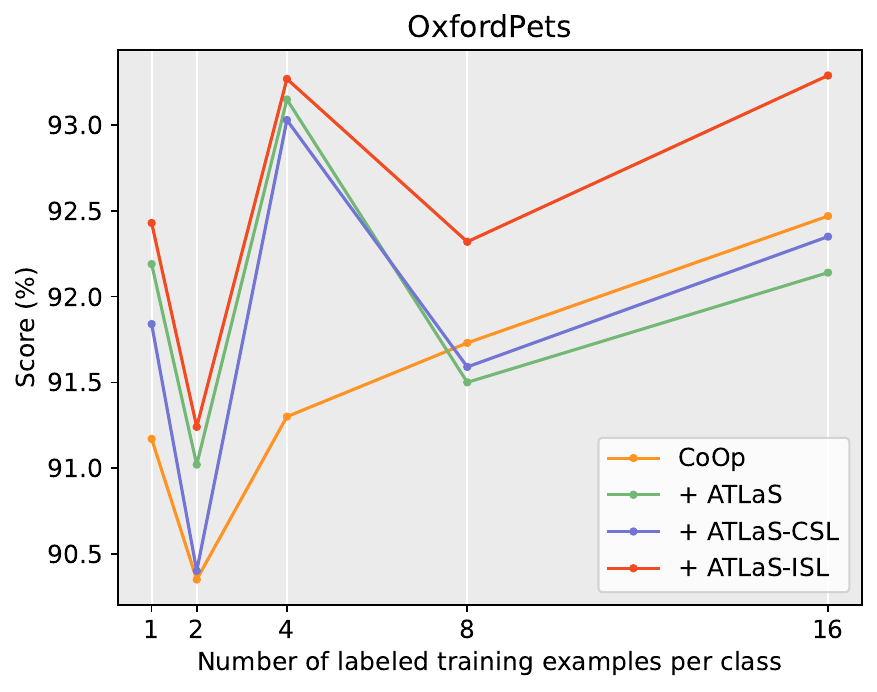}
  }
\hfill
    \subfloat{
    \includegraphics[width=0.29\linewidth]{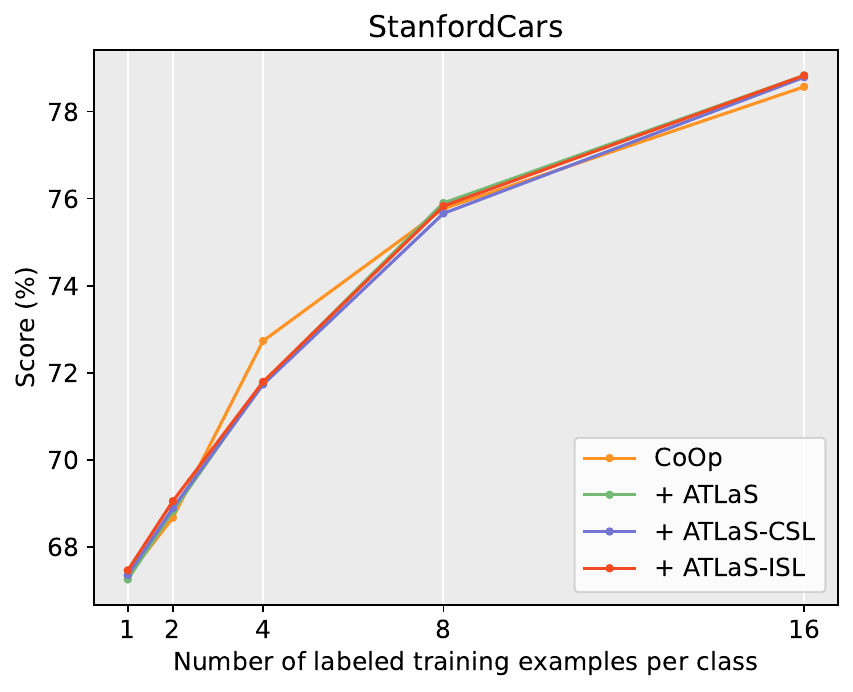}
  }
  \hfill
      \subfloat{
    \includegraphics[width=0.29\linewidth]{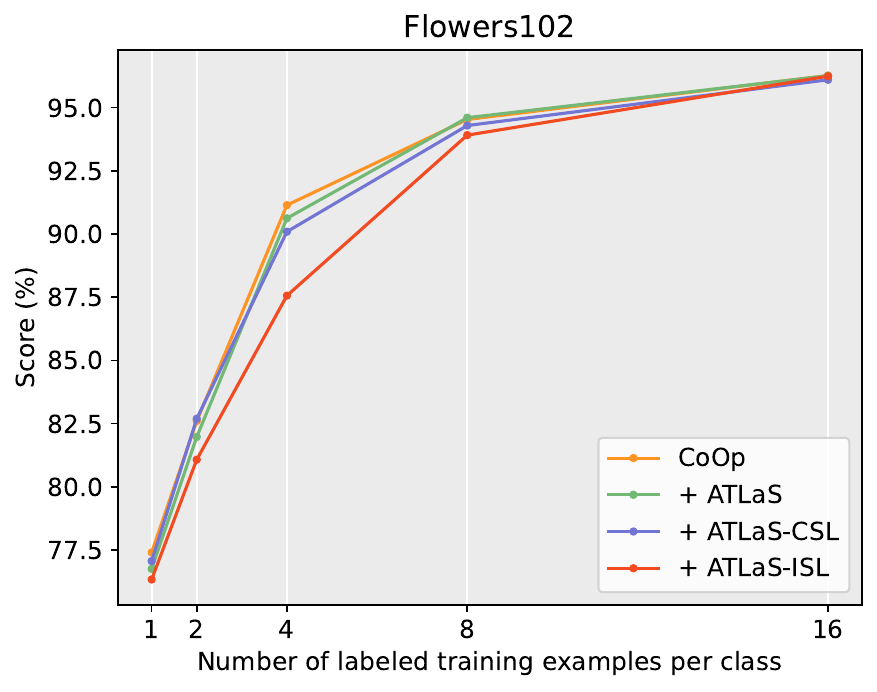}
  }

    \subfloat{
    \includegraphics[width=0.29\linewidth]{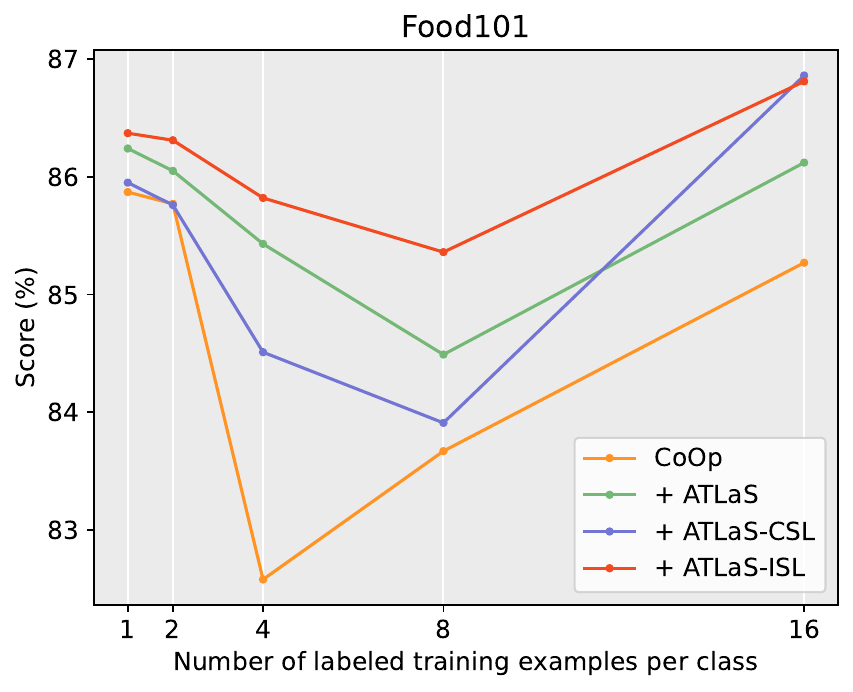}
  }
\hfill
    \subfloat{
    \includegraphics[width=0.29\linewidth]{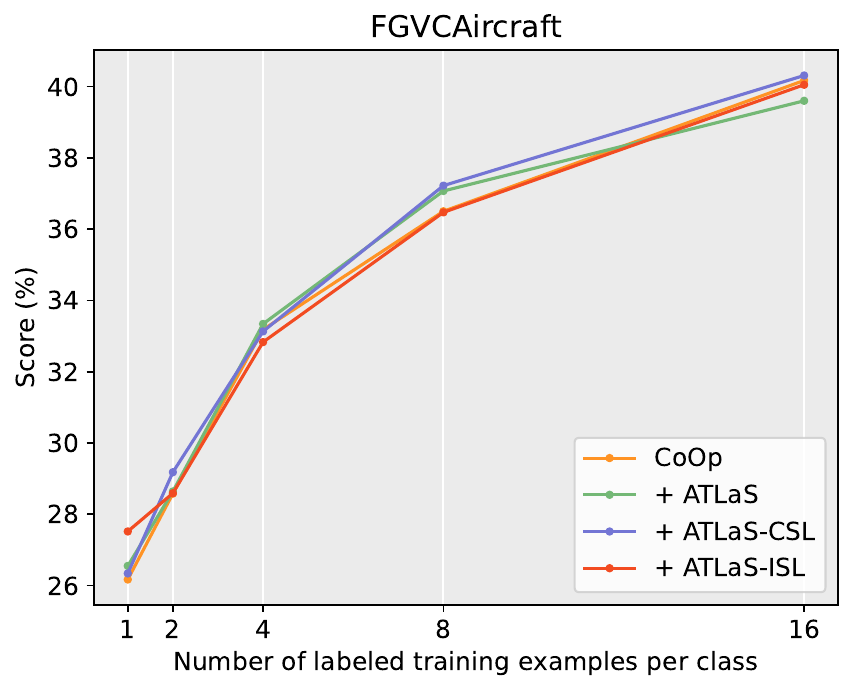}
  }
  \hfill
      \subfloat{
    \includegraphics[width=0.29\linewidth]{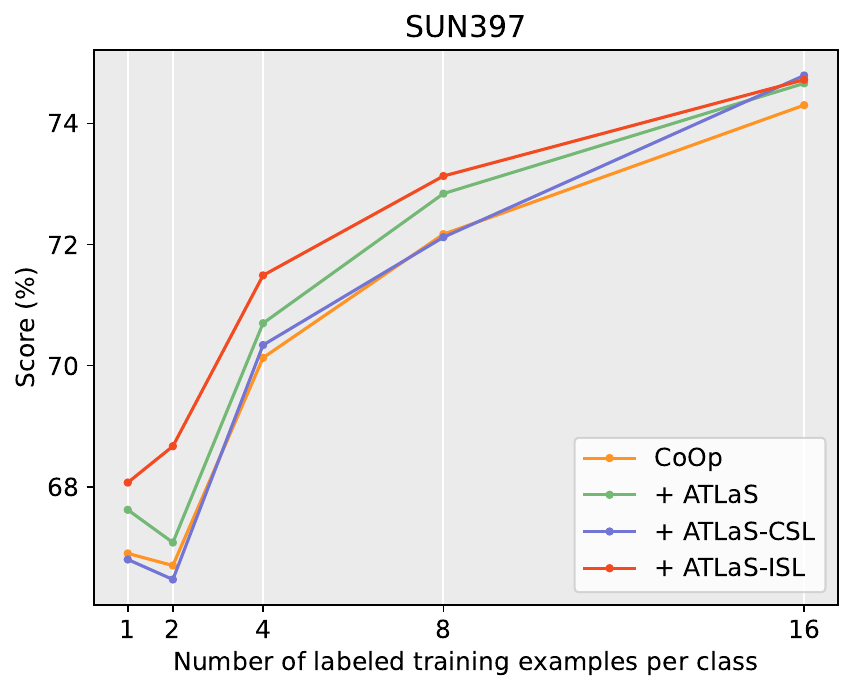}
  }

      \subfloat{
    \includegraphics[width=0.29\linewidth]{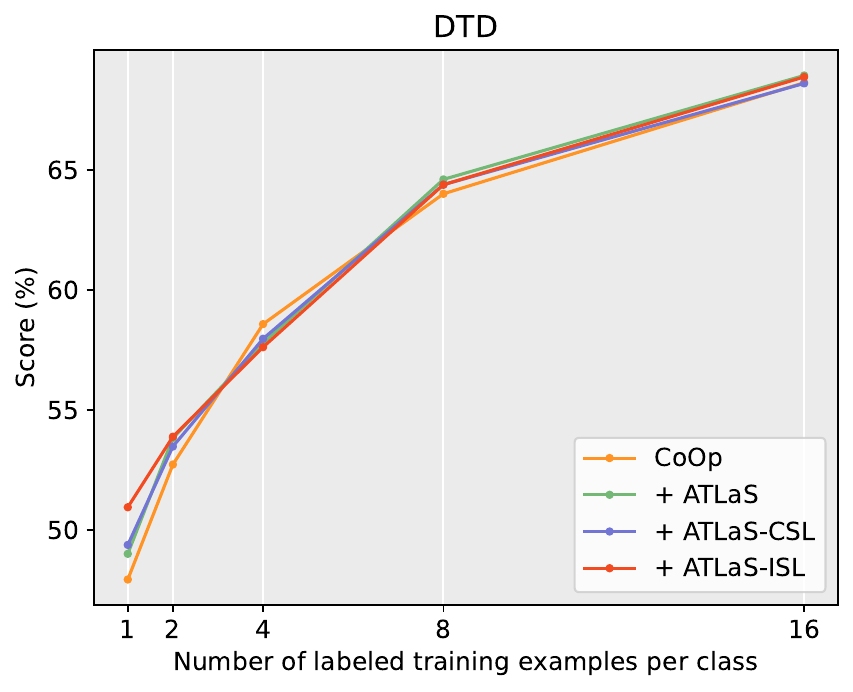}
  }
\hfill
    \subfloat{
    \includegraphics[width=0.29\linewidth]{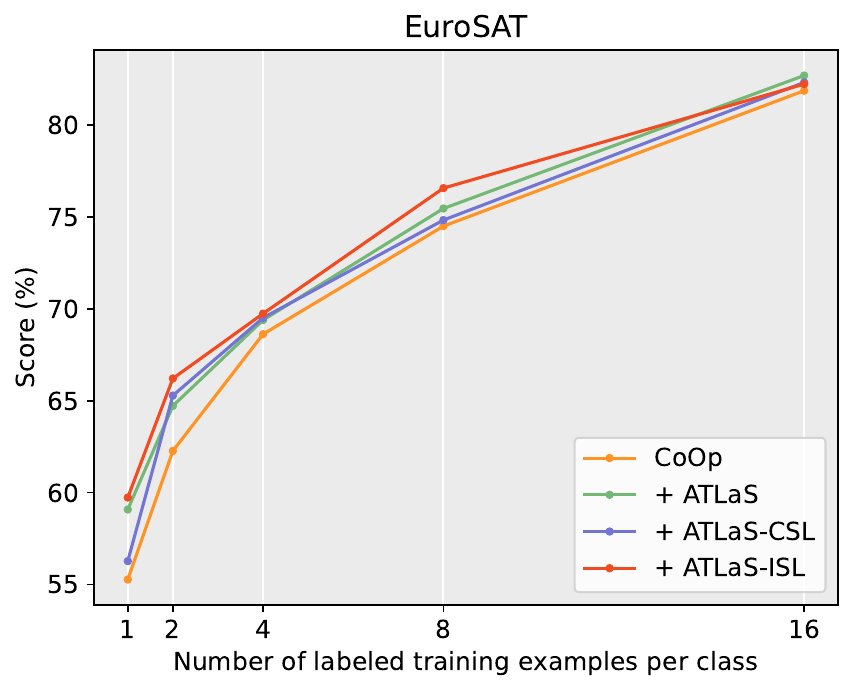}
  }
  \hfill
      \subfloat{
    \includegraphics[width=0.29\linewidth]{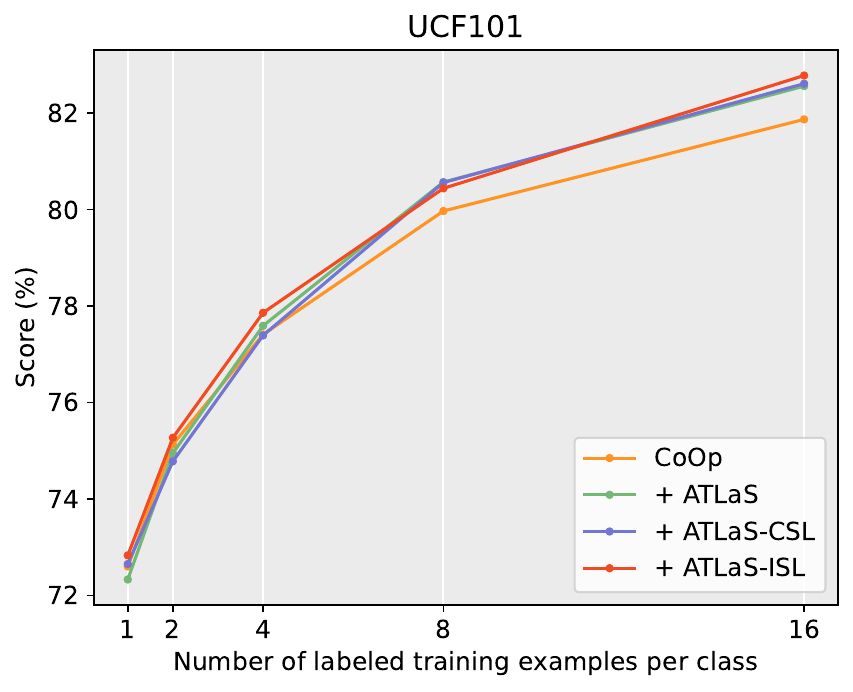}
  }

  \caption{Performance of various methods on the 11 datasets under the few-shot classification setting.}
  \label{fig_few_shot}
\end{figure*}

In the few-shot classification setting, prompts are trained using a limited number of labeled samples and evaluated on the remaining samples. We conduct few-shot classification using 1, 2, 4, 8, and 16 samples per class. The indicator function $\delta$ is always set to 1 under the few-shot classification setting. As visualized in \cref{fig_few_shot}, the advantages of the proposed method are pronounced even with very few training examples. ATLaS and its variants consistently outperform the CoOp baseline across all shots in terms of the average performance on the 11 datasets.
Those results demonstrate that our method's ability to enhance generalization is not contingent on large amounts of training data, making it highly valuable for practical, low-resource applications.

\subsection{Ablation Studies}
\label{sec:ablation}
In this section, we conduct ablation studies to analyze the effectiveness of the alternating mechanism and the proposed soft labels (i.e., CSL and ISL) in the proposed ATLaS method. The experiments are performed using CoOp \cite{coop} as the baseline method under the base-to-new generalization setting. 

\cref{tab:ablation} shows the average performance of CoOp equipped with different LS strategies on the 11 datasets. Based on the results, we have several observations.
\begin{itemize}
\item 
CoOp with vanilla LS degrades the performance of CoOp (i.e., -0.45\% in the base-class accuracy, -0.91\% in the new-class accuracy, -0.75\% in the harmonic mean accuracy), suggesting that such a naive combination makes the inconsistency of labels used in the pre-training and fine-tuning phases, weakening generalization performance.
In contrast, CoOp with ATLaS, which alternately uses one-hot and soft labels to supervise prompt tuning, outperforms CoOp by 1.96\% in the new-class accuracy (i.e., 65.19\% vs. 63.23\%) and 1.14\% in the harmonic mean accuracy (i.e., 72.80\% vs. 71.66\%), demonstrating that such alternating fine-tuning can enhance prompt generalization.
\item
Supervising prompt tuning with CSL and ISL as soft labels improves the new-class accuracy (i.e., +2.5\% for CSL, +10.41\% for ISL) and harmonic mean accuracy (i.e., +0.64\% for CSL, 3.58\% for ISL) at the expense of decreasing the base-class accuracy (i.e., -2.36\% for CSL, -5.79\% for ISL). This occurs because both CSL and ISL leverage the inherent properties of CLIP, allowing them to enhance prompt generalization while preserving the inherent knowledge of CLIP. Building upon this foundation, the alternating mechanism further improves their performance, that is, compared with CoOp, ATLaS-CSL and ATLaS-ISL improve the new-class accuracy by +4.82\% and +10.39\% respectively, and harmonic mean accuracy by +2.73\% and +5.13\%. Those results indicate that both the alternating mechanism and soft labels can enhance prompt generalization.
\item
Alternating optimization outperforms joint optimization when using both one-hot labels and soft labels to train prompts.
That is, CoOp with ATLaS outperforms CoOp with `vanilla LS + y' in terms of the harmonic mean accuracy (i.e., 72.80\% v.s. 71.68\%), further demonstrating the effectiveness of the proposed alternating mechanism.
\end{itemize}

\begin{table}[t]
  \centering
  \caption{Ablation study for the designed soft labels and alternating mechanism. The average performance of 11 datasets is reported. Prompts are trained with 16 instances per base class. `+ $\mathbf{y}$' denotes co-optimization with one-hot label $\mathbf{y}$, e.g., `CSL + $\mathbf{y}$' is to jointly optimize the prompts using two cross-entropy losses constructed by CSL and the one-hot label $\mathbf{y}$ in each iteration. The number in the parentheses denotes the improvement or degradation over the baseline method.}
  \resizebox{\linewidth}{!}{
  \setlength{\tabcolsep}{1.3mm}{
    \begin{tabular}{l|c|ccc}
    \toprule
    \multicolumn{1}{c|}{\multirow{2}[4]{*}{Setting}} & \multirow{2}[4]{*}{Alternating} & \multicolumn{3}{c}{Avg. acc. over 11 datasets (\%)} \\
\cmidrule{3-5}          &       & Base  & New   & H \\
    \midrule
    CoOp (Baseline) & \textcolor{gray}{\XSolidBrush} & 82.69 & 63.23 & 71.66 \\
    \midrule
      + vanilla LS & \textcolor{gray}{\XSolidBrush} & 82.24 & 62.32 & 70.91 (\textcolor{red}{-0.75}) \\
      + vanilla LS + $\mathbf{y}$ & \textcolor{gray}{\XSolidBrush} & 82.27  & 63.51  & 71.68 (+0.02) \\
    \rowcolor{Gray}   + ATLaS & \Checkmark & 82.42 & 65.19 & \textbf{72.80} (\textcolor{blue}{+\textbf{1.14}}) \\
    \midrule
      + CSL & \textcolor{gray}{\XSolidBrush} & 80.33 & 65.73 & 72.3 (+0.64) \\
      + CSL + $\mathbf{y}$ & \textcolor{gray}{\XSolidBrush} & 82.25  & 65.17  & 72.72 (+1.06) \\
    \rowcolor{Gray}   + ATLaS-CSL & \Checkmark & 82.03 & 68.05 & \textbf{74.39} (\textcolor{blue}{+\textbf{2.73}}) \\
    \midrule
      + ISL & \textcolor{gray}{\XSolidBrush} & 76.90 & 73.64 & 75.24 (+3.58) \\
      + ISL+ $\mathbf{y}$ & \textcolor{gray}{\XSolidBrush}  & 81.00  & 71.90  & 76.18 (+4.52) \\
    \rowcolor{Gray}   + ATLaS-ISL & \Checkmark & 80.25 & 73.62 & \textbf{76.79} (\textcolor{blue}{+\textbf{5.13}}) \\
    \bottomrule
    \end{tabular}}}
  \label{tab:ablation}
\end{table}%

\begin{figure*}[tb]
\subfloat[]
{\includegraphics[width=0.3\linewidth]{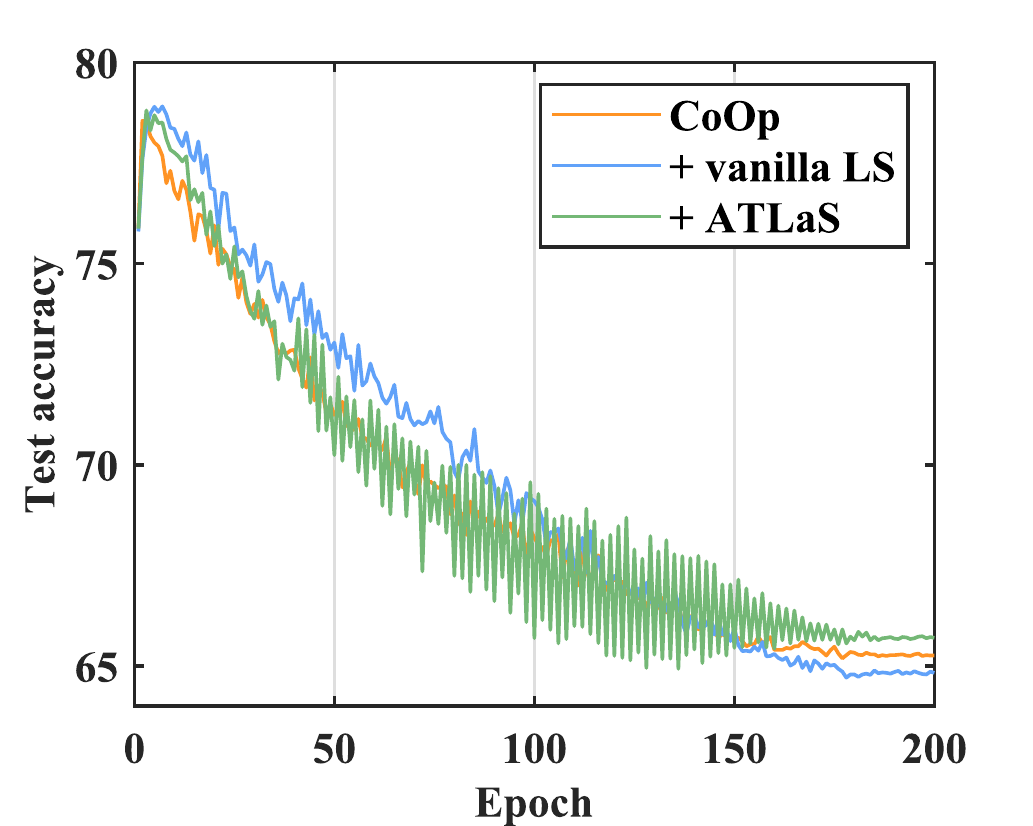}
    \label{sun_1}
}
  \hfill
\subfloat[]{
    \includegraphics[width=0.3\linewidth]{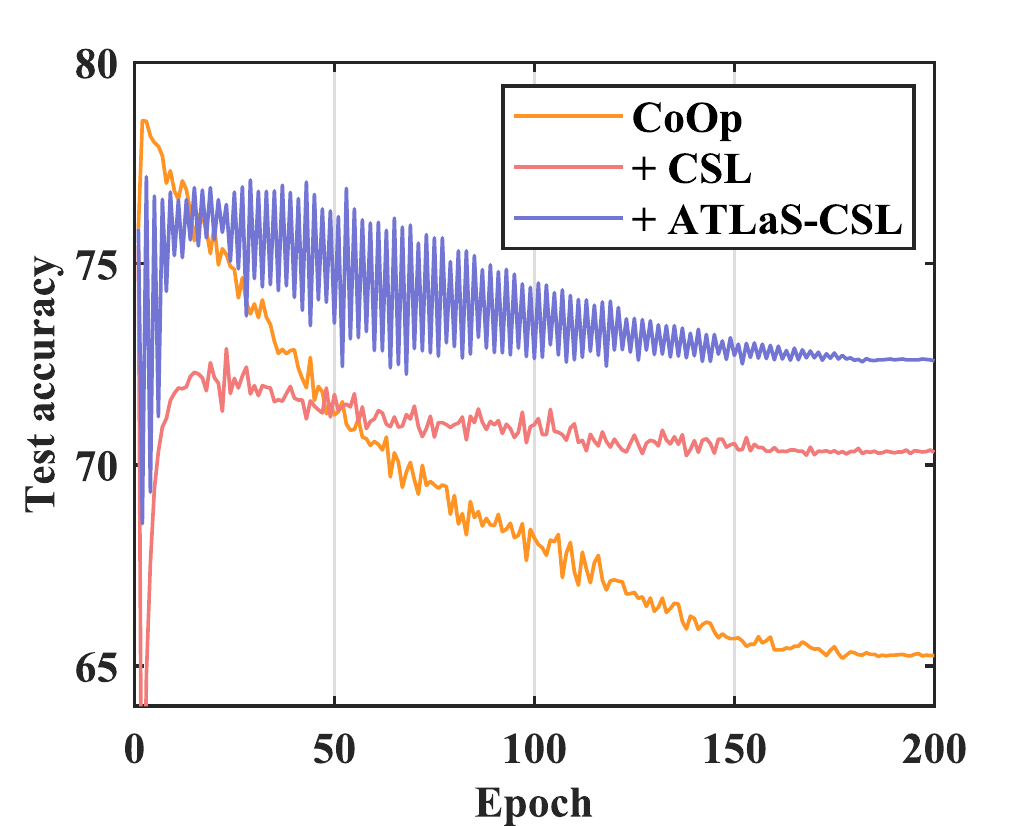}
    \label{sun_2}
}
  \hfill
\subfloat[]{
    \includegraphics[width=0.3\linewidth]{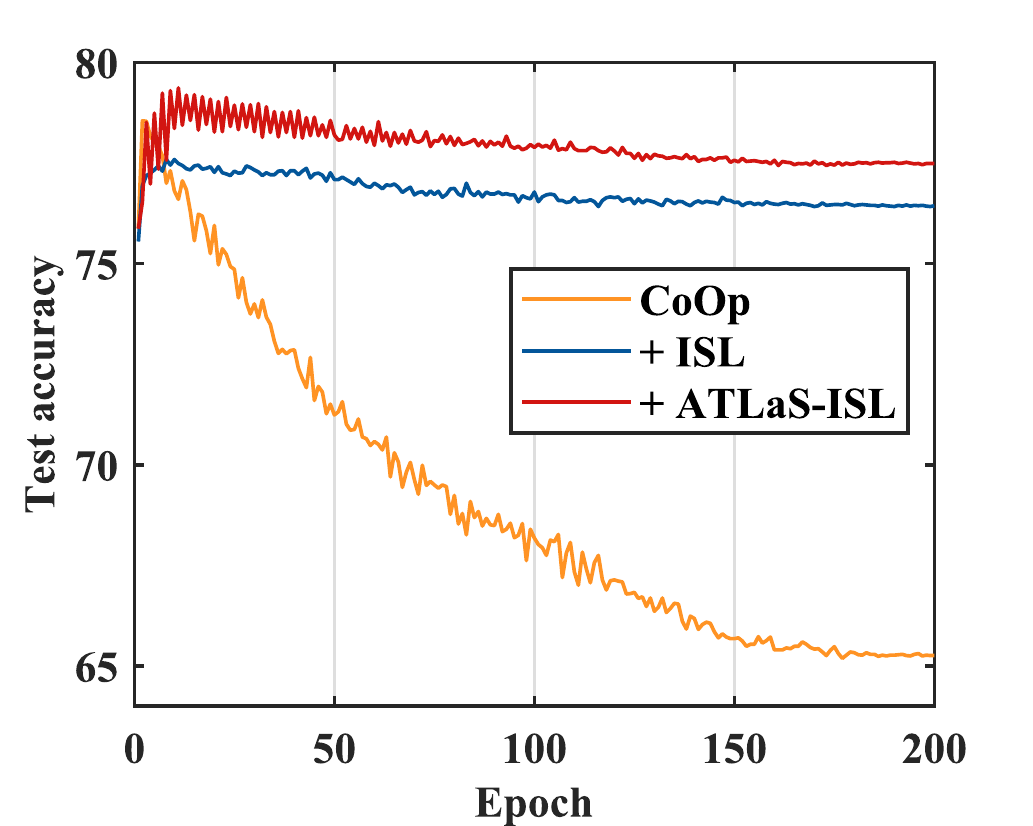}
    \label{sun_3}
}
  \caption{Comparison of test accuracies for CoOp with different LS strategies on new classes of SUN397.}
  \label{fig_sun397}
\end{figure*}

\cref{fig_sun397} shows the test accuracy of CoOp and CoOp with vanilla LS, ATLaS, CSL, ATLaS-CSL, ISL, and ATLaS-ISL on new classes of the SUN397 dataset over the training process. The test accuracy of CoOp first increases and then drops sharply. Integrating CoOp with vanilla LS does not improve the test accuracy while employing the alternating mechanism yields improvement. CoOp with both CSL and ISL shows higher test accuracy than CoOp and converges earlier. Building upon this, CoOp with ATLaS-CSL and ATLaS-ISL outperforms CoOp with CSL and ISL, respectively. This improved performance suggests that the consistency of labels used in the pre-training and fine-tuning phases can lead to better generalization.

\subsection{Sensitivity Analysis}
\begin{figure*}[htbp]
  \centering
  \subfloat[]{
    \includegraphics[width=0.22\linewidth]{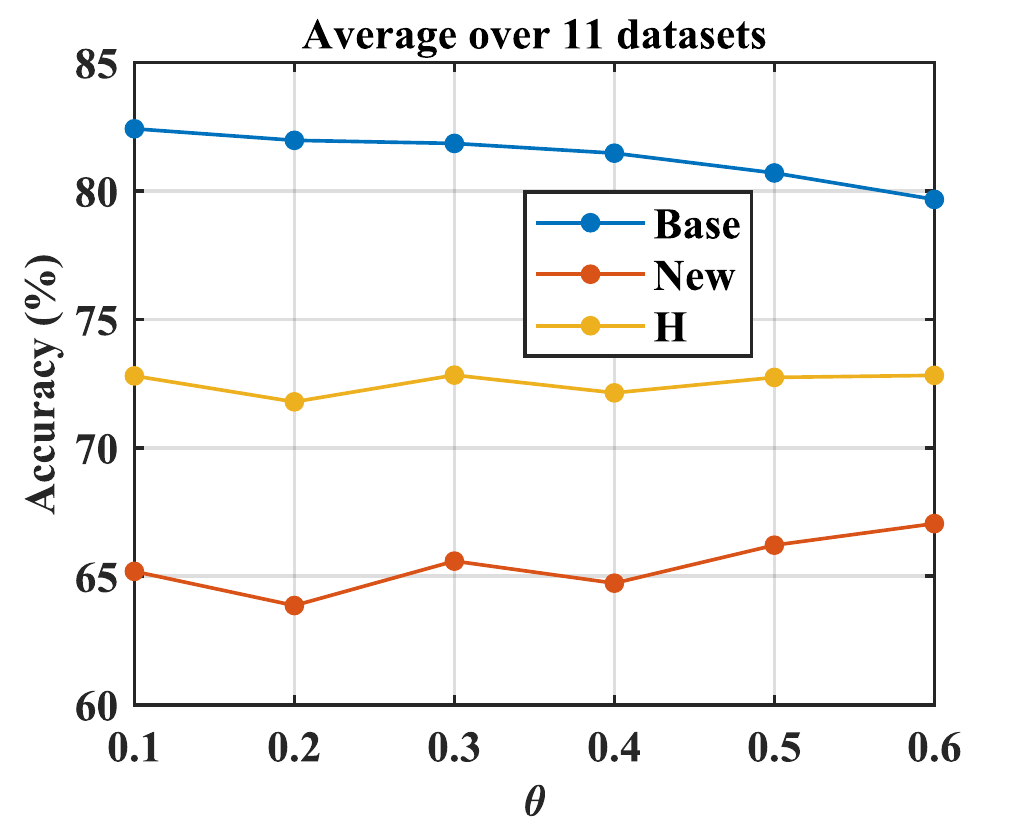}
    \label{theta}
}
  \hfill
  \subfloat[]{
    \includegraphics[width=0.22\linewidth]{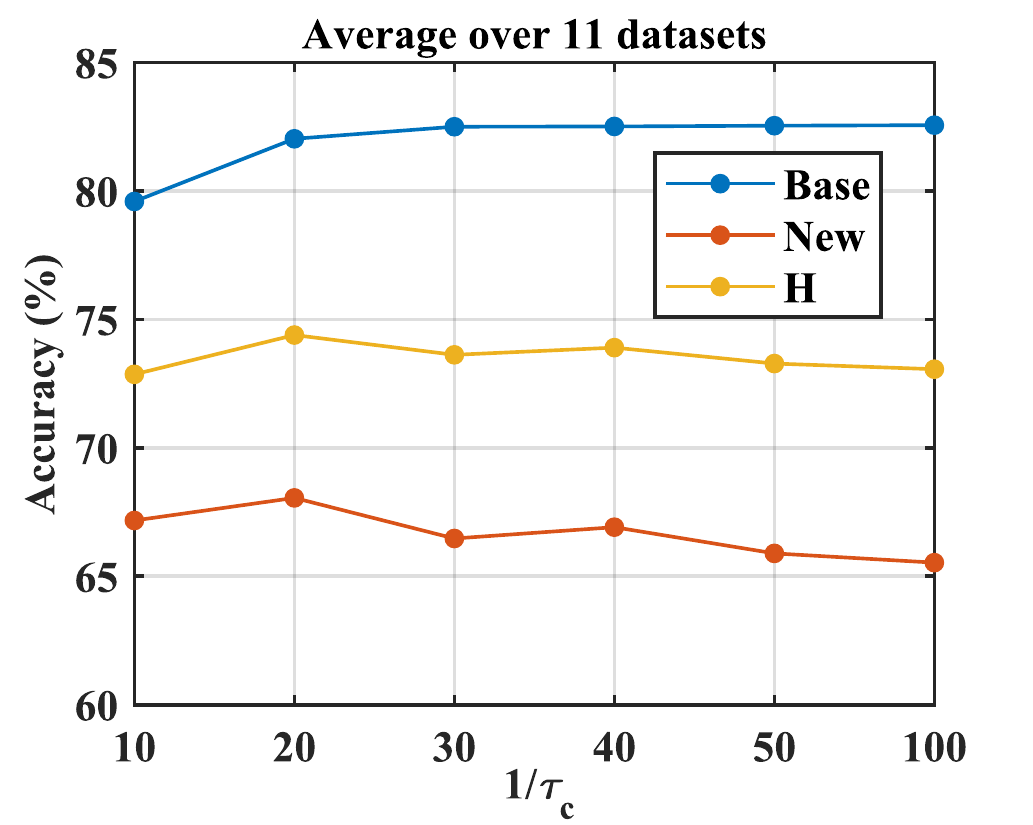}
    \label{tau}
}
  \hfill
  \subfloat[]{
    \includegraphics[width=0.22\linewidth]{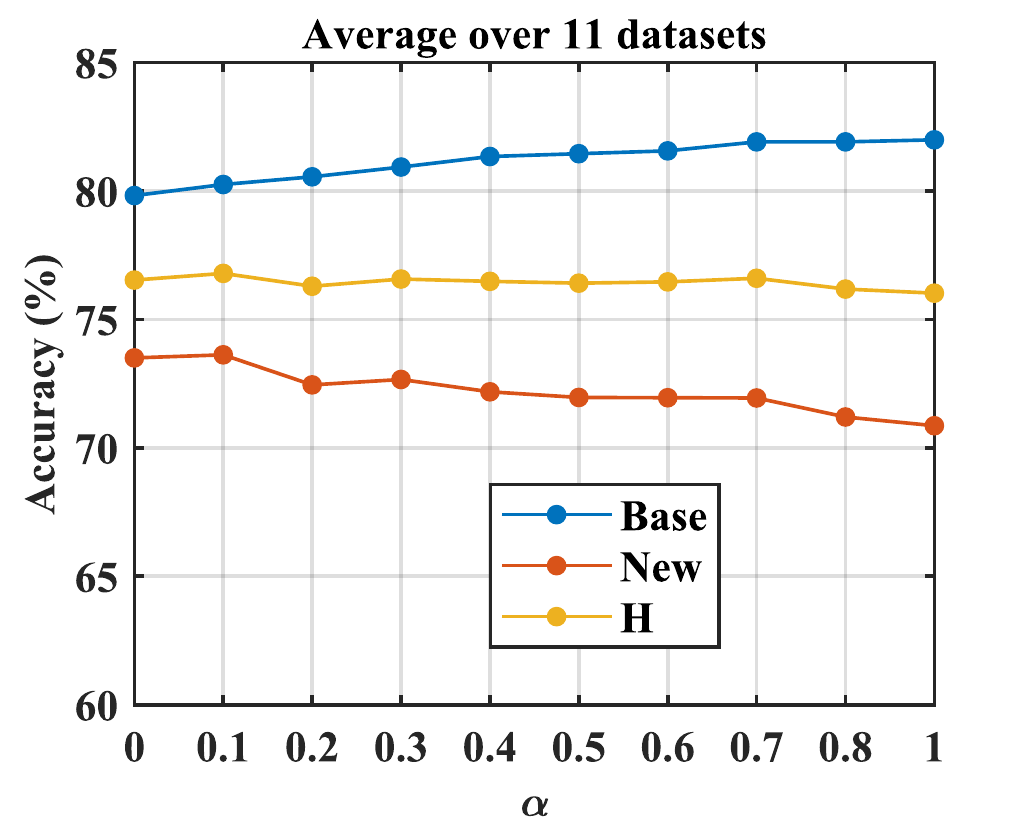}
    \label{alpha}
}
\hfill
  \subfloat[]{
    \includegraphics[width=0.22\linewidth]{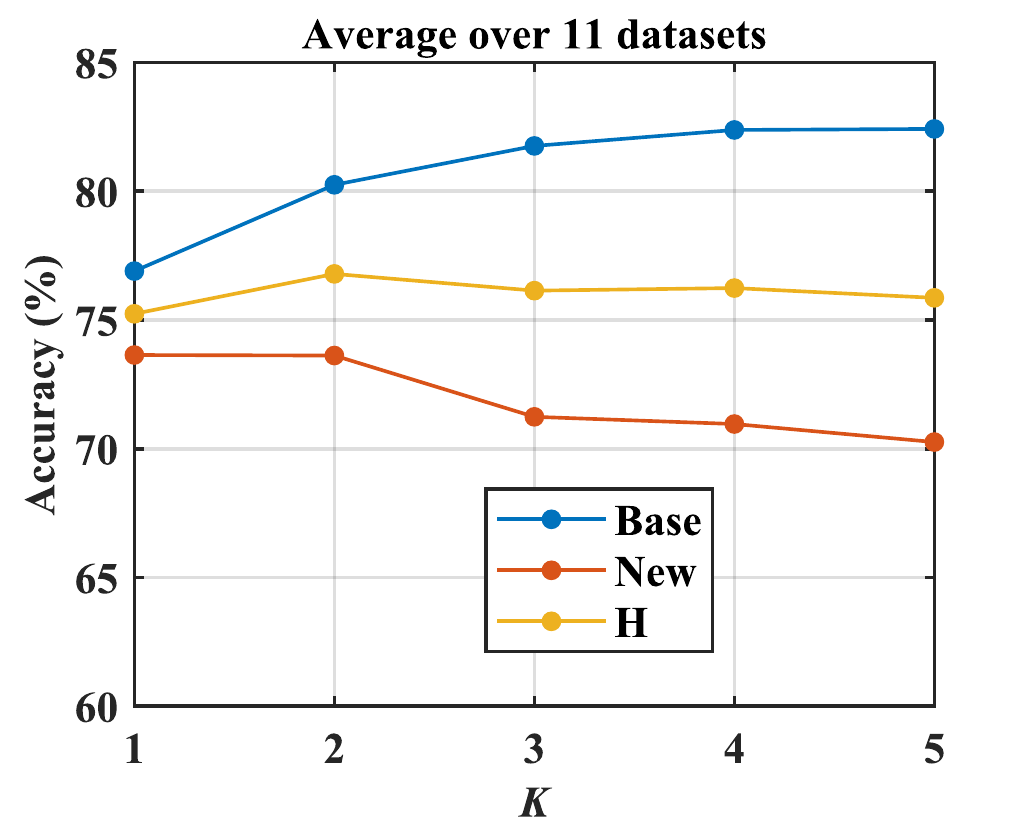}
    \label{K}
}
  \caption{The average performance of ATLaS (a), ATLaS-CSL (b), ATLaS-ISL (c, d) on all datasets with respect to hyper-parameters (i.e., $\theta$, $\tau _c$, $\alpha$, and $K$) under the base-to-new generalization setting.}
  \label{fig_parameter}

\end{figure*}

In this section, we investigate the impact of four hyper-parameters (i.e., $\theta$, $\tau _c$, $\alpha$, and $K$) on the average performance across 11 datasets under the base-to-new generalization setting and the results are shown in \cref{fig_parameter}.

\noindent{\textbf{Impact of the smoothing parameter $\theta$.}}
The smoothing parameter $\theta$ in Eq.~(\ref{eq_ls}) balances one-hot labels against the uniform distribution. \cref{theta} shows the average performance of CoOp with \textbf{ATLaS} across the 11 datasets for $\theta$ values in \{0.1,\ldots,0.6\}. As $\theta$ increases, soft labels exhibit reduced confidence in target classes and increased attention to non-target classes. This leads to a gradual decline in the base-class accuracy and improvement in the new-class accuracy, while the harmonic mean accuracy remains relatively stable.

\noindent{\textbf{Impact of the temperature parameter $\tau _c$.}}
The temperature parameter $\tau _c$ in Eq.~(\ref{eq_a_ij}) controls the smoothness of the similarity distribution across inter-class textual embeddings. A lower $\tau _c$ sharpens the distribution, enhancing sensitivity to inter-class distinctions, while a higher value smooths the distribution. \cref{tau} shows the average performance of CoOp with \textbf{ATLaS-CSL} across 11 datasets for $1/\tau _c$ values in \{10, 20,\ldots, 100\}. 
It can be seen that as $\tau _c$ decreases, the base-class accuracy initially increases and subsequently stabilizes, while both the new-class accuracy and the harmonic mean accuracy exhibit an initial increase followed by a decline.

\noindent{\textbf{Impact of the rectification coefficient $\alpha$.}}
The rectification coefficient $\alpha$ in Eq.~(\ref{eq_ISL}) controls how significantly the one-hot label modifies the soft label.
Higher $\alpha$ values intensify the rectification applied to the prediction probabilities of CLIP.
\cref{alpha} shows the average performance of CoOp with \textbf{ATLaS-ISL} across 11 datasets for $\theta$ values in \{0.1, 0.2,\ldots, 1\}.
As $\alpha$ increases, the base-class accuracy gradually improves, while the new-class accuracy initially increases, then declines gently. The harmonic mean accuracy remains generally stable across this range.

\noindent{\textbf{Impact of the alternating period $K$.}}
The alternating period $K$ regulates the ratio between epochs using one-hot labels and soft labels for prompt tuning. \cref{K} shows the average performance of CoOp with \textbf{ATLaS-ISL} across 11 datasets for $K$ in \{1, 2,\ldots, 5\}. Note that when $K=1$, the alternating training mechanism is disabled, and only instance-wise soft labels are used for training. As $K$ increases, the base-class accuracy gradually improves, while the new-class accuracy decreases, which aligns with our expectation since a larger $K$ will increase the proportion of one-hot labels used during training. Notably, the harmonic mean accuracy with the alternating mechanism (i.e., $K>1$) consistently outperforms that without the alternating mechanism (i.e., $K=1$), validating the effectiveness of the proposed alternating mechanism.

Overall, the results indicate that the performance of ATLaS, ATLaS-CSL, and ATLaS-ISL is not so sensitive to the values of hyper-parameters, making setting those hyper-parameters easier.

\subsection{Combination of CSL and ISL}
\label{appendix:combination}
\begin{figure}[!tbph]
  \centering
  \includegraphics[width=0.65\linewidth]{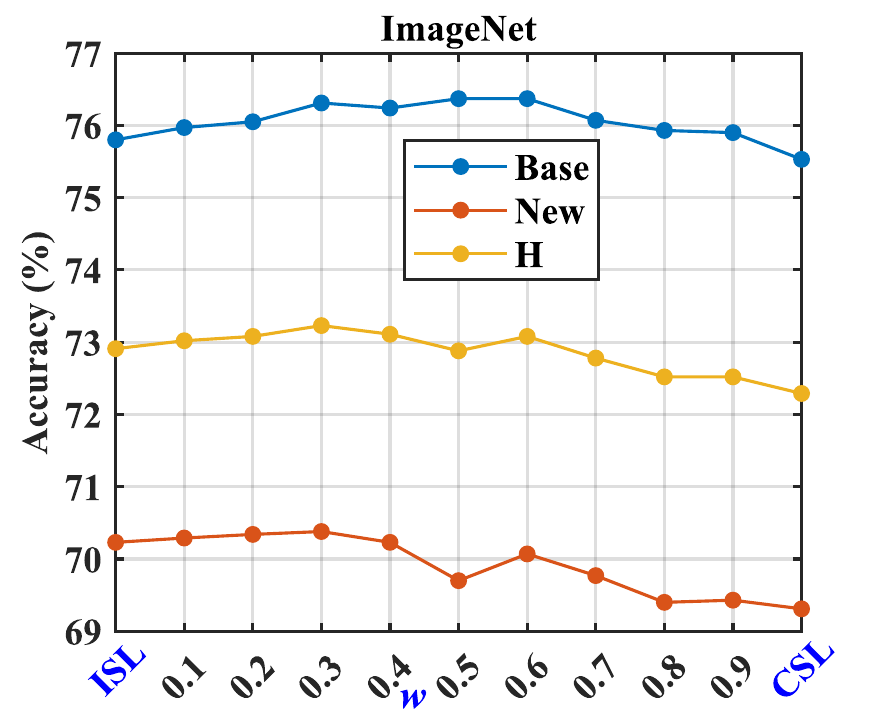}
   \caption{ATLaS with mixed ISL and CSL on the ImageNet dataset under the base-to-new generalization setting.}
   \label{fig_ISL_CSL}
\end{figure}

We explored CSL and ISL mixing ($w\times \mathrm{CSL}+\left( 1-w \right) \times \mathrm{ISL}$). As shown in \cref{fig_ISL_CSL}, on ImageNet, ATLaS achieves the best performance at $w=0.3$, surpassing individual CSL and ISL. However, we found that this mixing approach does not outperform standalone ISL or CSL across all scenarios.

\subsection{Further Analysis of the Alternating Period $K$}
\label{appendix:k}
\begin{figure}[!tbph]
  \centering
  \includegraphics[width=0.65\linewidth]{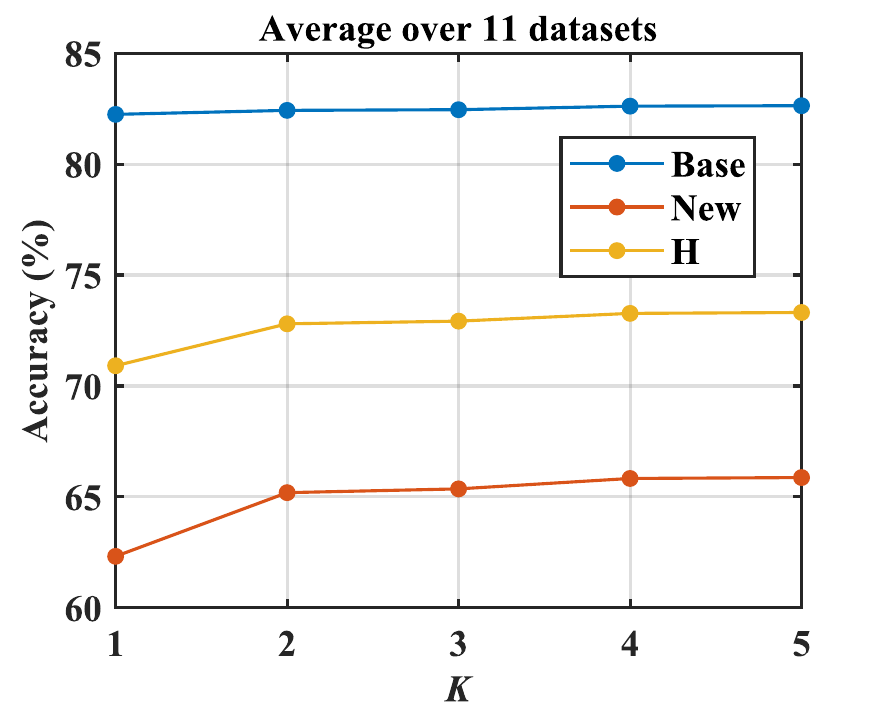}
   \caption{The average performance of ATLaS on all datasets with different $K$ under the base-to-new
generalization setting}
   \label{fig_K_ATLaS}
\end{figure}

As mentioned in Sec. \ref{subsec:analysis}, vanilla LS uses a uniform distribution to generate low-quality soft labels ($\kappa >1$). In such case, the alternating mechanism can improve the performance of vanilla LS. Figure \ref{fig_K_ATLaS} shows that when the alternating training mechanism is used (i.e., $K>1$), ATLaS maintains advantages over vanilla LS.

However, when high-quality soft labels (e.g., ISL) are used, $\kappa$ can be less than 1. Under this condition, Theorem \ref{theorem_1} indicates that non-alternating LS achieves a tighter convergence bound than ATLaS.
Therefore, Figure \ref{K} also aligns with our theoretical analysis, where $K=1$ (using ISL without alternating training) achieves better new-class accuracy compared to other $K$ using the alternating mechanism.
Nevertheless, the alternating mechanism can enhance the overall performance. Figure \ref{K} shows that the highest H score, which reflects balanced effectiveness on both Base and New classes, is achieved at $K=2$. This occurs because ATLaS-ISL improves base-class accuracy through one-hot label training while preventing overfitting to base classes.

\subsection{Computational Cost}
\label{appendix:cost}

We have explored the strong flexibility and effectiveness of ATLaS. In this section, we will examine the additional computational cost introduced by ATLaS, which mainly comes from generating CSL and ISL. According to \cref{tabcost}, generating CSL and ISL is computationally efficient, taking only 1.96s and 2.47s (merely 0.11$\%$ and 0.14$\%$ of the 1746s training time) on ImageNet. As one-time preprocessing steps, their costs are negligible given their benefits.

\begin{table}[htbp]
    \centering
  \caption{Computational cost of the proposed ATLaS, where the model is trained on the base classes of ImageNet with 16-shot examples per class.}

    \begin{tabular}{lllc}
    \toprule
    Method & Train. time & Infer.time & H (Avg.) \\
    \midrule
    CoOp  & 1746s & 2.33ms & 71.66 \\
    +ATLaS & +0s   & +0ms  & 72.30 \\
    +ATLaS-CSL & +1.96s & +0ms  & 74.39 \\
    +ATLaS-ISL & +2.47s & +0ms  & 76.79 \\
    \bottomrule
    \end{tabular}
  \label{tabcost}%
\end{table}

%% file: 5_conclusion.tex
\section{Conclusion}
\label{sec:conclusion}
In this work, we propose a novel ATLaS framework, which alternates between using standard one-hot labels and soft labels,  for prompt tuning with LS. First, we address the ineffectiveness of vanilla LS in prompt tuning through ATLaS. 
Second, we introduce two types of efficient offline soft labels that further enhance prompt tuning performance when combined with ATLaS. 
Third, the convergence properties of ATLaS are analyzed, showing its advantages over the vanilla LS. 
Finally, extensive experiments across the 11 datasets demonstrate that ATLaS can be effectively integrated with existing prompt tuning methods. In our future work, we will investigate the extension of ATLaS to other applications.


%% file: 6_appendix.tex
\renewcommand{\thetable}{A\arabic{table}}
\renewcommand{\thefigure}{A\arabic{figure}}

\noindent \textbf{\large{Contents}}
\begin{enumerate}
    \item Appendix \ref{appedix:proof} — Proof of Lemma
            \begin{itemize}
            \item \ref{appendix:lemma} — Proofs
            \item \ref{appendix:lemma2} — Technical Lemma \ref{lemma:2}
            \item \ref{appendix:proof} — Proof of Theorem \ref{theorem_1}
            \item \ref{appendix_Convergence} — Convergence Analysis for Label Smoothing
            \end{itemize}
    \item Appendix \ref{appendix:imp_detail} — Implementation Details
    \item Appendix \ref{appendix:exp_detail} — Experimental Details
            \begin{itemize}
            \item \ref{appendix:base2new} — Full Results of the Base-to-New Generalization Setting
            \end{itemize}

    
\end{enumerate}

\setcounter{section}{0}
\section{Proofs}
\label{appedix:proof}
In the following proofs, when the variable \(\mathbf{v}\) for which a gradient is to be taken is obvious, we use \(\nabla\) to substitute \(\nabla_{\mathbf{v}}\) for simplicity.

\subsection{Proof of Lemma \ref{lemma:1}}
\label{appendix:lemma}

Recall Lemma \ref{lemma:1} in Section \ref{sec:method}:
\setcounter{lemma}{0}
\begin{lemma}
    \[
    \mathbb{E}_{(\mathbf{x}, \mathbf{\hat{y}})} \left\| \nabla F(\mathbf{v}) - \nabla  \ell(\mathbf{\hat{y}}, f(\mathbf{v}; \mathbf{x})) \right\|^2 \leq \hat{\sigma}^2 = \kappa \sigma^2
    \]
    where \(\kappa > 0\) is a constant, and \(\sigma^2\) is the variance described in Assumption \ref{assump:2}.
\end{lemma}

\textit{Proof.} 

With labels \(\hat{\mathbf{y}} = \frac{1}{C}\mathbf{1}_C\) and considering the cross-entropy loss, it follows that:
\[
\nabla \ell(\hat{\mathbf{y}}, f(\mathbf{v}; \mathbf{x})) = f(\mathbf{v}; \mathbf{x}) - \hat{\mathbf{y}} = f(\mathbf{v}; \mathbf{x}) - \frac{1}{C}\mathbf{1}_C
\]
Thus, we have:
\setcounter{equation}{5}
\begin{align}
    & \mathbb{E}_{(\mathbf{x}, \mathbf{\hat{y}})} \left\| \nabla F(\mathbf{v}) - \nabla \ell(\mathbf{\hat{y}}, f(\mathbf{v}; \mathbf{x})) \right\|^2 \notag \\
    & = \mathbb{E}_{(\mathbf{x},\mathbf{\hat{y}})} \left\| \nabla F(\mathbf{v}) - f(\mathbf{v}; \mathbf{x}) + \frac{1}{C}\mathbf{1}_C \right\|^2
\label{appendix_eq_1}
\end{align}

Given Assumption \ref{assump:2}, the variance of the stochastic gradient with one-hot labels \(\mathbf{y}\) is:
\begin{align}
& \mathbb{E}_{(\mathbf{x}, \mathbf{y})} \left\| \nabla F(\mathbf{v}) - \nabla \ell(\mathbf{y}, f(\mathbf{v}; \mathbf{x})) \right\|^2 \notag \\
& = \mathbb{E}_{(\mathbf{x}, \mathbf{y})} \left\| \nabla F(\mathbf{v}) - f(\mathbf{v}; \mathbf{x}) + \mathbf{y} \right\|^2 \leq \sigma^2.
\label{appendix_eq_2}
\end{align}

Based on Eq.~(\ref{appendix_eq_2}), Eq.~(\ref{appendix_eq_1}) should also be bounded by some value \(\hat{\sigma}^2\). we also introduce a scaling factor \(\kappa = \frac{\hat{\sigma}^2}{\sigma^2}\) such that \(\hat{\sigma}^2 = \kappa \sigma^2\).

This completes the proof.
\subsection{Technical Lemma \ref{lemma:2}}
\label{appendix:lemma2}
Recall the optimization target is to minimize $F(\mathbf{v})$, formulated as:
\begin{equation}\mathop{\mathrm{min}}_{\mathbf{v}} F(\mathbf{v}) \coloneqq \mathbb{E}_{(\mathbf{x},\mathbf{y})}[\ell(\mathbf{y},f(\mathbf{v};\mathbf{x}))],
\label{fml:1}    
\end{equation}
 where the cross-entropy loss function is\[\ell \left( \mathbf{y},f\left( \mathbf{v};\mathbf{x} \right) \right) =-\sum_{i=1}^C{y_i\log \frac{\exp\mathrm{(}f_{i}^{\mathrm{s}}\left( \mathbf{v};\mathbf{x} \right) /\tau )}{\sum\nolimits_{j=1}^C{\exp\mathrm{(}f_{j}^{\mathrm{s}}\left( \mathbf{v};\mathbf{x} \right) /\tau )}}},\]
Let $p_i(\mathbf{v};\mathbf{x}) = -\log \frac{\exp\mathrm{(}f_{i}^{\mathrm{s}}\left( \mathbf{v};\mathbf{x} \right) /\tau )}{\sum\nolimits_{j=1}^C{\exp\mathrm{(}f_{j}^{\mathrm{s}}\left( \mathbf{v};\mathbf{x} \right) /\tau )}}$, \(p(\mathbf{v};\mathbf{x}) = \{p_1(\mathbf{v};\mathbf{x}),...,p_C(\mathbf{v};\mathbf{x})\} \in \mathbb{R}^C\), then the optimization target in (1) can be rewritten as follow, where \(\langle \cdot,\cdot\rangle\) means inner product:
\begin{equation}
    \mathop{\mathrm{min}}_{\mathbf{v}} F(\mathbf{v}) \coloneqq \mathbb{E}_{(\mathbf{x},\mathbf{y})}[\langle\mathbf{y}, p(\mathbf{v};\mathbf{x})\rangle]. 
\label{appendix_eq_4}
\end{equation}
Then the stochastic gradient with respective to $\mathbf{v}$ is:
\begin{equation}
    \nabla \ell \left( \mathbf{y},f\left( \mathbf{v};\mathbf{x} \right) \right) =\langle\mathbf{y}, \nabla p(\mathbf{v};\mathbf{x})\rangle
\end{equation}
Let $\xi \in \{0,1\}$ be an indicator of whether the label-smoothed label or the original one-hot label is used at iteration $t$ (i.e., $\xi = 1$ for using the one-hot label \(\mathbf{y}_t\), and $0$
for using the label-smoothed label $\mathbf{y}_t^{\mathrm{LS}}$). Therefore, the label used in ATLaS at iteration $t$ can be formulated as:
\begin{align}
    \mathbf{y}_t^{\mathrm{ATLaS}} &= \xi \mathbf{y}_t + (1-\xi) \mathbf{y}_t^{\mathrm{LS}} \notag \\
    &\mathop{=}^{(a)} (1-\theta)\mathbf{y}_t + \xi\theta \mathbf{y}_t + (1-\xi)\theta \hat{\mathbf{y}}_t,
\end{align}
where $(a)$ uses \(\mathbf{y}_t^{\mathrm{LS}} = (1-\theta)\mathbf{y}_t + \theta \hat{\mathbf{y}}_t\).
\begin{lemma}\label{lemma:2}
Under Assumptions \ref{assump:1} and \ref{assump:2}, the following inequation holds:
\begin{align}
    \mathbb{E}\left[\|\nabla \ell(\mathbf{y}_t^{\mathrm{ATLaS}}, f(\mathbf{v}_t;\mathbf{x}_t))-\nabla F(\mathbf{v}_t) \|^2\right] & \notag \\ 
    \leq (1-\frac{\theta}{K}+\frac{\theta\kappa}{K})\sigma^2 & 
\end{align}
\end{lemma}

\textit{Proof.} Combing the Eq.~(\ref{fml:1}) and Eq.~(\ref{appendix_eq_4}), we have:
\begin{align}
\nabla\ell(\mathbf{y}_t^{\mathrm{ATLaS}}, f(\mathbf{v}_t;\mathbf{x}_t)) \notag & 
    = (1-\theta)\nabla\ell(\mathbf{y}_t, f(\mathbf{v}_t;\mathbf{x}_t)) 
    \\ \notag & 
    + \xi\theta \nabla\ell(\mathbf{y}_t,f(\mathbf{v}_t;\mathbf{x}_t))
    \\  & 
    + (1-\xi)\theta \nabla\ell(\hat{\mathbf{y}}_t, f(\mathbf{v}_t;\mathbf{x}_t)).
\label{appendix_eq_8}
\end{align}
Therefore,
\begin{align}
    \notag & \mathbb{E}\left[\|\nabla \ell(\mathbf{y}_t^{\mathrm{ATLaS}}, f(\mathbf{v}_t;\mathbf{x}_t))-\nabla F(\mathbf{v}_t) \|^2\right] \notag 
    \\ \notag & 
    = \mathbb{E}\left[\|
    (1-\theta)[\nabla\ell(\mathbf{y}_t, f(\mathbf{v}_t;\mathbf{x}_t))-\nabla F(\mathbf{v}_t)] + 
    \right. 
    \\ \notag & 
    \quad\quad\quad\quad\quad \quad\quad
    \xi\theta [\nabla\ell(\mathbf{y}_t,f(\mathbf{v}_t;\mathbf{x}_t))-\nabla F(\mathbf{v}_t)] +
    \\ \notag & 
    \quad\quad\quad\quad\quad\quad\quad \left.
    (1-\xi)\theta [\nabla\ell(\hat{\mathbf{y}}_t, f(\mathbf{v}_t;\mathbf{x}_t))-\nabla F(\mathbf{v}_t)]
    \|^2 \right], \
    \\ \notag & 
    \mathop{=}^{(a)} p(\xi=1)\mathbb{E}\left[\|\nabla\ell(\mathbf{y}_t,f(\mathbf{v}_t;\mathbf{x}_t))-\nabla F(\mathbf{v}_t)\|^2\right] 
    \\ \notag & 
    \quad + p(\xi=0)\mathbb{E}\Big[\|(1-\theta)[\nabla\ell(\mathbf{y}_t, f(\mathbf{v}_t;\mathbf{x}_t))-\nabla F(\mathbf{v}_t)]  + 
    \\ \notag & \quad\quad\quad\quad\quad\quad\quad \theta [\nabla\ell(\hat{\mathbf{y}}_t,f(\mathbf{v}_t;\mathbf{x}_t))-\nabla F(\mathbf{v}_t)]\|^2\Big] 
    \\ \notag & 
    \mathop{\leq}^{(b)} p(\xi=1)\mathbb{E}\left[\|\nabla\ell(\mathbf{y}_t,f(\mathbf{v}_t;\mathbf{x}_t))-\nabla F(\mathbf{v}_t)\|^2\right] 
    \\ \notag & 
    \quad + p(\xi=0)\Big((1-\theta)\mathbb{E}\left[\|[\nabla\ell(\mathbf{y}_t, f(\mathbf{v}_t;\mathbf{x}_t))-\nabla F(\mathbf{v}_t)]\|^2\right] 
    \\ \notag & \quad\quad\quad\quad\quad\quad\quad + \theta\mathbb{E}\left[\|[\nabla\ell(\mathbf{\hat{y}}_t, f(\mathbf{v}_t;\mathbf{x}_t))-\nabla F(\mathbf{v}_t)]\|^2\right]\Big) \\ \notag
    & \mathop{=}^{(c)} \frac{K-1}{K}\sigma^2 + \frac{1-\theta + \theta\kappa}{K}\sigma^2
    \\ &
    = (1-\frac{\theta}{K}+\frac{\theta\kappa}{K})\sigma^2
\end{align}

\noindent where (a) uses \(\xi \in \{0,1\}\); (b) applies the convexity of the norm: \(\|(1 - \theta)\mathbf{a} + \theta \mathbf{b}\|^2 \leq (1 - \theta)\| \mathbf{a}\|^2 + \theta\| \mathbf{b}\|^2\); (c) considers the ATLaS alternating frequency and label frequencies: one-hot label frequency \(p(\xi=1)=\frac{K-1}{K}\) and smoothed label frequency \(p(\xi=0)=\frac{1}{K}\), along with the variances \(\mathbb{E}_{(\mathbf{x}_t,\mathbf{y}_t)}\|\nabla F(\mathbf{v}_t)-\nabla\ell(\mathbf{y}_t,f(\mathbf{v}_t;\mathbf{x}_t))\|^2 = \sigma^2\) and \(\mathbb{E}_{(\mathbf{x}_t,\mathbf{\hat{y}}_t)}\|\nabla F(\mathbf{v}_t)-\nabla\ell(\mathbf{\hat{y}}_t,f(\mathbf{v}_t;\mathbf{x}_t))\|^2 = \kappa\sigma^2\).

\subsection{Proof of Theorem \ref{theorem_1}}
\label{appendix:proof}
Here we first restate Theorem \ref{theorem_1} in Section \ref{sec:method} as follows:

\setcounter{theorem}{0}
\begin{theorem}

With Assumptions \ref{assump:1} and \ref{assump:2}, 
learning rate 
$\eta = \frac{1}{\beta}$ 
and 
label smoothing parameter
$\theta$, ATLaS satisfies
\[ \frac{1}{T}\sum_{t=0}^{T-1}
\mathbb{E}[\| \nabla_{\mathbf{v_t}} F(\mathbf{v}_t) \|^2] \leq \frac{2F(\mathbf{v}_0)}{\eta T} +\left(1-\frac{\theta}{K}+\frac{\theta\kappa}{K}\right)\sigma^2,
\]
where $T$ is the total number of iterations in the optimization process and $\mathbf{v}_t$ is the prompt vectors at iteration $t$.
\end{theorem}
\textit{Proof.} Recall Assumption \ref{assump:1} says the objective function $F(\mathbf{v})$ is $\beta$-smooth, i.e., $\left|\nabla F(\mathbf{v})-\nabla F(\mathbf{u})\left| \leq \beta\right|\mathbf{v}-\mathbf{u}\right|$, which has an equivalent expression \cite{nesterov2013introductory}:
\begin{equation}
    F(\mathbf{v})-F(\mathbf{u}) \leq \langle\nabla F(\mathbf{u}), \mathbf{v-\mathbf{u}}\rangle + \frac{\beta}{2}\|\mathbf{v}-\mathbf{u}\|^2.
\label{appendix_eq_10}
\end{equation}
Therefore,
\begin{align}
\notag & F(\mathbf{v}_{t+1})-F(\mathbf{v}_t) \\ \notag
& \leq \langle\nabla F(\mathbf{v}_t), \mathbf{v_{t+1}}-\mathbf{v}_t\rangle + \frac{\beta}{2}\|\mathbf{v}_{t+1}-\mathbf{v}_t\|^2 \\ \notag
& \mathop{=}^{(a)} -\eta\langle\nabla F(\mathbf{v}_t), \nabla \ell(\mathbf{y}_{t}^{\mathrm{ATLaS}}, f(\mathbf{v}_t;\mathbf{x}_t))\rangle \\ \notag & \quad +\frac{\eta^2\beta}{2}\| \nabla \ell(\mathbf{y}_{t}^{\mathrm{ATLaS}}, f(\mathbf{v}_t;\mathbf{x}_t))\|^2 
\\ \notag  & \mathop{=}^{(b)} 
-\frac{\eta}{2}\|\nabla F(\mathbf{v}_t)\|^2
+ \frac{\eta}{2}\| \nabla F(\mathbf{v}_t)-\nabla \ell(\mathbf{y}_{t}^{\mathrm{ATLaS}}, f(\mathbf{v}_t;\mathbf{x}_t))\|^2 
\\ \notag &
\quad + \underbrace{\frac{\eta(\eta\beta-1)}{2}\|\nabla \ell(\mathbf{y}_{t}^{\mathrm{ATLaS}}, f(\mathbf{v}_t;\mathbf{x}_t))\|^2}_{=0}
\\ &
\mathop{=}^{(c)} 
-\frac{\eta}{2}\|\nabla F(\mathbf{v}_t)\|^2
+ \frac{\eta}{2}\| \nabla F(\mathbf{v}_t)-\nabla \ell(\mathbf{y}_{t}^{\mathrm{ATLaS}}, f(\mathbf{v}_t;\mathbf{x}_t))\|^2, 
\end{align}
where (a) use the update of $\mathbf{v}_{t+1}$; (b) is due to $\langle\mathbf{a},-\mathbf{b}\rangle = \frac{1}{2}(\|\mathbf{a}-\mathbf{b}\|^2)-\|\mathbf{a}\|^2-\|\mathbf{b}\|^2$; (c) is because \(\eta=\frac{1}{\beta}\).

Taking the expectation of both sides in the above inequation, we obtain:
\begin{align}
\notag & \mathbb{E}[F(\mathbf{v}_{t+1})-F(\mathbf{v}_t)]
\\ \notag & \leq -\frac{\eta}{2}\mathbb{E}\left[\|\nabla F(\mathbf{v}_t)\|^2\right]  
\\ \notag & 
\quad+ \frac{\eta}{2} \mathbb{E}\left[\|\nabla F(\mathbf{v}_t)-\nabla \ell(\mathbf{y}_{t}^{\mathrm{ATLaS}}, f(\mathbf{v}_t;\mathbf{x}_t))\|^2\right]
\\ & \leq -\frac{\eta}{2}\mathbb{E}\left[\|\nabla F(\mathbf{v}_t)\|^2\right] 
+ \frac{\eta}{2} \left(1-\frac{\theta}{K}+ \frac{ \theta\kappa}{K}\right)\sigma^2, 
\end{align}
where the last inequality is due to Lemma \ref{lemma:2}. Then we have,
\begin{align}
 \notag &
    \frac{1}{T}\sum_{t=0}^{T-1} \mathbb{E}[\nabla \|F(\mathbf{v}_t)\|^2]
\\ \notag &
    \mathop{\leq}^{(a)} \frac{2F(\mathbf{v}_0)}{\eta T} + \left(1-\frac{\theta}{K} + \frac{\theta\kappa}{K}\right)\sigma^2,
\end{align}
where (a) is derived by summing both sides of Eq.~(\ref{appendix_eq_10}) and rearranging, 
We thus finish the proof.

\subsection{Convergence Analysis for Label Smoothing}
\label{appendix_Convergence}

In this section, we include the convergence analysis for label smoothing derived by \cite{xu2020towards}. Note here the context vector \(\mathbf{v}_t\) are the only learnable parameters, slightly different from the version in \cite{xu2020towards} where all the parameters $\mathbf{x}_t$ in the model are learnable. 

\begin{lemma}\label{lemma:3}
Under Assumption \ref{assump:1} and \ref{assump:2}, the following inequation hold:
\begin{align}
    \mathbb{E}\left[\|\nabla \ell(\mathbf{y}_t^{\mathrm{LS}}, f(\mathbf{v}_t;\mathbf{x}_t))-\nabla F(\mathbf{v}_t) \|^2\right] & \notag \\ 
    \leq (1-\theta+\theta\kappa)\sigma^2 &
\end{align}
\end{lemma}

\textit{Proof.} Combine $\mathbf{y}_t^{\mathrm{LS}} = (1-\theta)\mathbf{y}_t+\theta\hat{\mathbf
{y}}_t$ and Eq.~(\ref{fml:1}), we have:
\begin{align}
    \nabla\ell(\mathbf{y}_t^{\mathrm{LS}},f(\mathbf{v}_t;\mathbf{x}_t)) \notag & = (1-\theta)\nabla(\ell(\mathbf{y}_t^{\mathrm{LS}},f(\mathbf{v}_t;\mathbf{x}_t)))  
    \\  &
    + \theta \ell(\hat{\mathbf{y}}_t^{\mathrm{LS}},f(\mathbf{v}_t;\mathbf{x}_t))
\label{appendix_eq_14}
\end{align}
Therefore,
\begin{align}
\notag & \mathbb{E}[\|\nabla\ell(\mathbf{y}_t^{\mathrm{LS}},f(\mathbf{v}_t;\mathbf{x}_t))\|^2] \\ \notag & 
= \mathbb{E}\left[\|(1-\theta)\nabla\ell(\mathbf{y}_t,f(\mathbf{v}_t;\mathbf{x}_t)) + \theta\nabla\ell(\hat{\mathbf{y}_t},f(\mathbf{v}_t;\mathbf{x}_t))\|^2\right]
\\ \notag &
\mathop{\leq}^{(a)} (1-\theta)\mathbb{E}\left[\|\nabla(\ell(\mathbf{y}_t,f(\mathbf{v}_t;\mathbf{x}_t))\|^2\right]  
\\ \notag &
\ \ \ \ \ \ \ \ \ \ \ + \theta\mathbb{E}\left[\|\nabla\ell(\hat{\mathbf{y}_t},f(\mathbf{v}_t;\mathbf{x}_t))\|^2\right]
\\ \notag &
\mathop{\leq}^{(b)} (1-\theta+\theta\kappa)\sigma^2,
\end{align}
where (a) applies the convexity of the norm:  \(\|(1 - \theta)\mathbf{a} + \theta \mathbf{b}\|^2 \leq (1 - \theta)\| \mathbf{a}\|^2 + \theta\| \mathbf{b}\|^2\); (b) uses the assumption \ref{assump:2} and Lemma \ref{lemma:1}, i.e., \(\mathbb{E}_{(\mathbf{x}_t,\mathbf{y}_t)}\|\nabla F(\mathbf{v}_t)-\nabla\ell(\mathbf{y}_t,f(\mathbf{v}_t;\mathbf{x}_t))\|^2 = \sigma^2\), and \((\mathbb{E}_{(\mathbf{x}_t,\mathbf{\hat{y}}_t)}\|\nabla F(\mathbf{v}_t)-\nabla\ell(\mathbf{\hat{y}}_t,f(\mathbf{v}_t;\mathbf{x}_t))\|^2 = \kappa\sigma^2\).

\begin{theorem}Under Assumption \ref{assump:1} and \ref{assump:2}, with $\eta = \frac{1}{\beta}$ 
, label smoothing satisfy:
\[
    \frac{1}{T}\sum_{0}^{T-1}
\mathbb{E}[\| \nabla F(\mathbf{v}_t) \|^2] \leq \frac{2F(\mathbf{v}_0)}{\eta T} + 
(1-\theta+\theta\kappa)\sigma^2
\]
\end{theorem}
\textit{Proof.} With Assumption \ref{assump:1}, we have Eq.~(\ref{appendix_eq_8}) and thus,

\begin{align}
\notag & F(\mathbf{v}_{t+1})-F(\mathbf{v}_t) \\ \notag
& \leq \langle\nabla F(\mathbf{v}_t), \mathbf{v_{t+1}}-\mathbf{v}_t\rangle + \frac{\beta}{2}\|\mathbf{v}_{t+1}-\mathbf{v}_t\|^2 \\ \notag
& \mathop{=}^{(a)} -\eta\langle\nabla F(\mathbf{v}_t), \nabla \ell(\mathbf{y}_{t}^{\mathrm{LS}}, f(\mathbf{v}_t;\mathbf{x}_t))\rangle \\ \notag & \quad +\frac{\eta^2\beta}{2}\| \nabla \ell(\mathbf{y}_{t}^{\mathrm{LS}}, f(\mathbf{v}_t;\mathbf{x}_t))\|^2 
\\ \notag  & \mathop{=}^{(b)} 
-\frac{\eta}{2}\|\nabla F(\mathbf{v}_t)\|^2
+ \frac{\eta}{2}\| \nabla F(\mathbf{v}_t)-\nabla \ell(\mathbf{y}_{t}^{\mathrm{LS}}, f(\mathbf{v}_t;\mathbf{x}_t))\|^2 
\\ \notag &
\quad + \underbrace{\frac{\eta(\eta\beta-1)}{2}\|\nabla \ell(\mathbf{y}_{t}^{\mathrm{LS}}, f(\mathbf{v}_t;\mathbf{x}_t))\|^2}_{=0}
\\ &
\mathop{=}^{(c)} 
-\frac{\eta}{2}\|\nabla F(\mathbf{v}_t)\|^2
+ \frac{\eta}{2}\| \nabla F(\mathbf{v}_t)-\nabla \ell(\mathbf{y}_{t}^{\mathrm{LS}}, f(\mathbf{v}_t;\mathbf{x}_t))\|^2, 
\end{align}
where (a) use the update of $\mathbf{v}_{t+1}$; (b) is due to $\langle\mathbf{a},-\mathbf{b}\rangle = \frac{1}{2}(\|\mathbf{a}-\mathbf{b}\|^2)-\|\mathbf{a}\|^2-\|\mathbf{b}\|^2$; (c) is because \(\eta=\frac{1}{\beta}\).

Taking the expectation of both sides in the above inequation, we obtain:
\begin{align}
\notag & \mathbb{E}[F(\mathbf{v}_{t+1})-F(\mathbf{v}_t)]
\\ \notag & \leq -\frac{\eta}{2}\mathbb{E}\left[\|\nabla F(\mathbf{v}_t)\|^2\right]  
\\ \notag & 
\quad+ \frac{\eta}{2} \mathbb{E}\left[\|\nabla F(\mathbf{v}_t)-\nabla \ell(\mathbf{y}_{t}^{\mathrm{LS}}, f(\mathbf{v}_t;\mathbf{x}_t))\|^2\right]
\\ & \leq -\frac{\eta}{2}\mathbb{E}\left[\|\nabla F(\mathbf{v}_t)\|^2\right] 
+ \frac{\eta}{2} \left((1-\theta+\theta\kappa)\sigma^2\right), 
\end{align}
where the last inequality is due to Lemma \ref{lemma:3}. Then we have,
\begin{align}
 \notag &
    \frac{1}{T}\sum_{t=0}^{T-1} \mathbb{E}[\nabla \|F(\mathbf{v}_t)\|^2]
\\ \notag &
    \mathop{\leq}^{(a)} \frac{2F(\mathbf{v}_0)}{\eta T} + (1-\theta+\theta\kappa)\sigma^2,
\end{align}
where (a) is derived by summing both sides of Eq.~(\ref{appendix_eq_14}) and rearranging, 
We thus finish the proof.
\label{appendix:proof_ls}

\section{Implementation Details}
\label{appendix:imp_detail}
All experiments are conducted based on NVIDIA GeForce RTX 3090. 

For the ATLaS method, the hyper-parameters ($K$, $\theta$, $\tau_c$, $\alpha$) are set to (2, 0.1, 0.05, 0.1) under the base-to-new generalization setting, (3, 0.05, 0.02, 0.1) under the few-shot classification setting, and (3, 0.1, 0.04, 1.0) under both the cross-dataset generalization and domain generalization settings, respectively.

The textual prompts for different datasets used in CSL and ISL follow \cite{clip, coop}, as shown below:
\begin{quote}
\begin{scriptsize}
\begin{verbatim}
ImageNet: "a photo of a [CLS]."
Caltech101: "a photo of a [CLS]." 
OxfordPets: "a photo of a [CLS], a type of pet." 
StanfordCars: "a photo of a [CLS]." 
OxfordFlowers: "a photo of a [CLS], a type of 
                 flower." 
Food101: "a photo of [CLS], a type of food." 
FGVCAircraft: "a photo of a [CLS], a type of 
                aircraft." 
SUN397: "a photo of a [CLS]." 
DTD: "a photo of a [CLS], a type of texture." 
EuroSAT: "a centered satellite photo of [CLS]." 
UCF101: "a photo of a person doing [CLS]." 
\end{verbatim}
\end{scriptsize}
\end{quote}

Note that [CLS] denotes the placeholder for the class name.


\section{Experimental Details}
\label{appendix:exp_detail}

Due to the page limit in the main body, we provide detailed results in this section.

\subsection{Full Results of the Base-to-New Generalization Setting}
\label{appendix:base2new}

In this section, we provide the base-to-new generalization performance on 11 datasets and additionally integrate our ATLaS and its variants with the state-of-the-art PromptSRC method \cite{promptsrc}. The results in \cref{appendix_tab:base2new_1,appendix_tab:base2new_2,appendix_tab:base2new_3} show that the proposed ATLaS method can consistently improve the new-class accuracy and harmonic mean accuracy over baseline methods, highlighting its effectiveness in enhancing prompt generalization. However, we observe limited performance gains when applying ATLaS to PromptSRC, potentially attributable to both methods leveraging CLIP's multi-modal capabilities to capture and exploit inter-class and instance-class relationships. Nonetheless, the proposed ATLaS method achieves performance improvements for PromptSRC on all the datasets.

\setcounter{table}{0}
\begin{table*}[htbp]
  \caption{The average performance on 11 datasets and the performance on the ImageNet, Caltech101, and OxfordPets datasets for various prompt tuning methods with or without ATLaS, ATLaS-CSL, and ATLaS-ISL under the base-to-new generalization setting. `H' denotes the harmonic mean accuracy. The best results are marked in bold.}
\resizebox{\linewidth}{!}{
\setlength{\tabcolsep}{4.7mm}{
  \centering
    \begin{tabular}{L{2cm}|ccc|ccc|ccc|ccc}
    \toprule
    \multicolumn{1}{c|}{\multirow{2}{*}{Methods}} & \multicolumn{3}{c|}{\textbf{Avg. over 11 datasets}} & \multicolumn{3}{c|}{ImageNet} & \multicolumn{3}{c|}{Caltech101} & \multicolumn{3}{c}{OxfordPets} \\
\cmidrule{2-13}          & Base  & New   & H     & Base  & New   & H     & Base  & New   & H     & Base  & New   & H \\
    \midrule
    CoOp  & 82.69  & 63.23  & 71.66  & 76.47  & 67.88  & 71.92  & 98.00  & 89.91  & 93.78  & 93.67  & 95.29  & 94.47  \\
    \midrule
      + ATLaS & 82.42  & 65.19  & 72.80  & 76.53  & 69.57  & 72.88  & 98.03  & 90.83  & 94.29  & 93.57  & 93.27  & 93.42  \\
      + ATLaS-CSL & 82.03  & 68.05  & 74.39  & 75.53  & 69.31  & 72.29  & 97.97  & 92.17  & 94.98  & 93.63  & 95.73  & 94.67  \\
   \cellcolor{Gray}+ ATLaS-ISL & \cellcolor{Gray}80.25  & \cellcolor{Gray}73.62  &\cellcolor{Gray}{\textbf{76.79}} &  75.80  &  70.23  & \textbf{72.91} &  97.67  &  94.63  &  \textbf{96.13} &  94.50  &  97.17  &  \textbf{95.82} \\
    \midrule
    CoCoOp & 80.47  & 71.69  & 75.83  & 75.98  & 70.43  & 73.10  & 97.96  & 93.81  & 95.84  & 95.20  & 97.69  & 96.43  \\
    \midrule
      + ATLaS & 80.13  & 73.43  & 76.64  & 76.20  & 70.57  & 73.28  & 97.73  & 94.50  & \textbf{96.09} & 95.47  & 97.50  & 96.47  \\
      + ATLaS-CSL & 79.79  & 73.69  & 76.62  & 76.20  & 70.63  & 73.31  & 97.70  & 93.97  & 95.80  & 95.23  & 97.77  & \textbf{96.48} \\
    \cellcolor{Gray}  + ATLaS-ISL & \cellcolor{Gray}79.19  & \cellcolor{Gray}74.37  & \cellcolor{Gray}{\textbf{76.70}} &  75.96  &  70.96  &  \textbf{73.37} &  97.74  &  94.03  &  95.85  &  94.33  &  96.96  &  95.63  \\
    \midrule
    DePT  & 83.56  & 71.92  & 77.30  & 77.19  & 70.24  & 73.55  & 98.35  & 94.03  & 96.14  & 94.56  & 97.58  & 96.05  \\
    \midrule
      + ATLaS & 83.81  & 73.67  & 78.42  & 77.18  & 70.20  & 73.52  & 98.43  & 94.72  & 96.54  & 94.90  & 97.54  & 96.20  \\
      + ATLaS-CSL & 83.76  & 73.85  & 78.49  & 77.27  & 70.53  & 73.75  & 98.32  & 95.01  & 96.64  & 94.88  & 97.54  & 96.19  \\
    \cellcolor{Gray}  + ATLaS-ISL & \cellcolor{Gray}83.80  & \cellcolor{Gray}75.05  & \cellcolor{Gray}{\textbf{79.18}} &  77.31  &  70.79  &  \textbf{73.91} &  98.45  &  95.45  &  \textbf{96.93} &  94.84  &  97.93  &  \textbf{96.36} \\
    \midrule
    IVLP  & 83.69  & 71.10  & 76.88  & 77.14  & 66.66  & 71.52  & 98.54  & 93.78  & 96.10  & 94.75  & 97.02  & 95.87  \\
    \midrule
      + ATLaS & 84.39  & 71.90  & 77.64  & 77.27  & 68.89  & 72.84  & 98.32  & 93.82  & 96.02  & 94.77  & 96.36  & 95.56  \\
   \cellcolor{Gray}+ ATLaS-CSL & \cellcolor{Gray}84.21  & \cellcolor{Gray}73.41  & \cellcolor{Gray}{\textbf{78.44}} &  77.17  &  67.02  &  71.74  &  98.43  &  94.25  &  96.29  &  94.99  &  97.33  &  96.15  \\
      + ATLaS-ISL & 84.26  & 73.06  & 78.26  & 77.48  & 69.26  & \textbf{73.14} & 98.56  & 94.18  & \textbf{96.32} & 95.87  & 97.26  & \textbf{96.56} \\
    \midrule
    MaPLe & 83.52  & 73.22  & 78.04  & 77.37  & 69.61  & 73.29  & 98.13  & 94.43  & 96.24  & 95.04  & 96.89  & 95.96  \\
    \midrule
      + ATLaS & 83.46  & 74.21  & 78.56  & 77.38  & 69.93  & 73.47  & 97.95  & 94.47  & 96.18  & 95.71  & 97.35  & 96.52  \\
      + ATLaS-CSL & 83.19  & 74.69  & 78.71  & 77.34  & 70.20  & 73.60  & 98.39  & 94.43  & \textbf{96.37} & 95.50  & 97.73  & 96.60  \\
    \cellcolor{Gray}+ ATLaS-ISL & \cellcolor{Gray}83.09  & \cellcolor{Gray}75.13  & \cellcolor{Gray}{\textbf{78.91}} &  77.45  &  70.74  &  \textbf{73.94} &  98.32  &  94.18  &  96.21  &  95.80  &  97.78  &  \textbf{96.78} \\
    \midrule
    PromptSRC & 84.23  & 75.73  & 79.76  & 77.58  & 70.58  & 73.91  & 98.08  & 94.00  & 96.00  & 95.23  & 97.28  & 96.24  \\
    \midrule
    \cellcolor{Gray}+ ATLaS & \cellcolor{Gray}84.14  & \cellcolor{Gray}76.25  & \cellcolor{Gray}{\textbf{80.00}} &  77.73  &  70.47  &  73.92  &  98.00  &  94.07  &  95.99  &  95.30  &  97.41  &  96.34  \\
      + ATLaS-CSL & 84.17  & 76.20  & 79.99  & 77.63  & 70.60  & \textbf{73.95} & 98.10  & 94.03  & 96.02  & 95.37  & 97.24  & 96.30  \\
      + ATLaS-ISL & 83.97  & 76.33  & 79.97 & 77.45  & 70.76  & \textbf{73.95} & 98.07  & 94.27  & \textbf{96.13} & 95.47  & 97.50  & \textbf{96.47} \\
    \bottomrule
    \end{tabular}}}
  \label{appendix_tab:base2new_1}%
\end{table*}%

\begin{table*}[htbp]
  \caption{The performance of various prompt tuning methods with or without ATLaS, ATLaS-CSL, and ATLaS-ISL on the StanfordCars, Flowers102, Food101, and FGVCAircraft datasets under the base-to-new generalization setting.}
\resizebox{\linewidth}{!}{
\setlength{\tabcolsep}{4.7mm}{
  \centering
    \begin{tabular}{L{2cm}|ccc|ccc|ccc|ccc}
    \toprule
    \multicolumn{1}{c|}{\multirow{2}{*}{Methods}} & \multicolumn{3}{c|}{StanfordCars} & \multicolumn{3}{c|}{Flowers102} & \multicolumn{3}{c|}{Food101} & \multicolumn{3}{c}{FGVCAircraft} \\
\cmidrule{2-13}          & Base  & New   & H     & Base  & New   & H     & Base  & New   & H     & Base  & New   & H \\
    \midrule
    CoOp  & 78.12  & 60.40  & 68.13  & 97.60  & 59.67  & 74.06  & 88.33  & 82.26  & 85.19  & 40.44  & 22.30  & 28.75  \\
    \midrule
      + ATLaS & 75.17  & 64.30  & 69.31  & 97.40  & 59.30  & 73.72  & 88.93  & 85.30  & 87.08  & 40.57  & 25.60  & 31.39  \\
      + ATLaS-CSL & 75.77  & 67.53  & 71.41  & 97.23  & 66.57  & 79.03  & 89.20  & 85.23  & 87.17  & 38.50  & 26.97  & 31.72  \\
     + ATLaS-ISL &  72.13  &  75.00  &  \textbf{73.54} &  91.10  &  74.27  &  \textbf{81.83} &  90.70  &  91.03  &  \textbf{90.86} &  37.47  &  32.10  &  \textbf{34.58} \\
    \midrule
    CoCoOp & 70.49  & 73.59  & \textbf{72.01} & 94.87  & 71.75  & \textbf{81.71} & 90.70  & 91.29  & 90.99  & 33.41  & 23.71  & 27.74  \\
    \midrule
      + ATLaS & 70.30  & 73.73  & 71.97  & 94.50  & 70.33  & 80.64  & 90.63  & 91.73  & \textbf{91.18} & 34.60  & 34.50  & 34.55  \\
      + ATLaS-CSL & 70.07  & 72.97  & 71.49  & 94.20  & 71.03  & 80.99  & 90.70  & 91.67  & \textbf{91.18} & 34.63  & 34.03  & 34.33  \\
     + ATLaS-ISL &  68.96  &  74.17  &  71.47  &  92.91  &  72.08  &  81.18  &  90.49  &  91.57  &  91.03  &  33.69  &  35.67  &  \textbf{34.65} \\
    \midrule
    DePT  & 79.35  & 72.43  & 75.73  & 98.04  & 71.84  & 82.92  & 90.43  & 91.45  & 90.94  & 41.92  & 22.45  & 29.24  \\
    \midrule
      + ATLaS & 79.41  & 71.15  & 75.05  & 98.26  & 73.88  & 84.34  & 90.37  & 91.47  & 90.92  & 43.02  & 32.31  & 36.90  \\
      + ATLaS-CSL & 79.49  & 71.91  & 75.51  & 98.26  & 75.18  & 85.18  & 90.32  & 91.44  & 90.88  & 43.38  & 32.15  & 36.93  \\
     + ATLaS-ISL &  79.37  &  72.74  &  \textbf{75.91} &  98.20  &  77.45  &  \textbf{86.60} &  90.48  &  91.68  &  \textbf{91.08} &  42.94  &  34.19  &  \textbf{38.07} \\
    \midrule
    IVLP  & 79.88  & 71.02  & 75.19  & 97.63  & 71.18  & 82.33  & 89.13  & 89.79  & 89.46  & 37.06  & 20.59  & 26.47  \\
    \midrule
      + ATLaS & 79.25  & 69.61  & 74.12  & 97.69  & 70.24  & 81.72  & 89.99  & 90.07  & 90.03  & 42.32  & 32.01  & 36.45  \\
     + ATLaS-CSL &  80.07  &  72.30  &  \textbf{75.99} &  97.85  &  73.19  &  \textbf{83.74} &  89.71  &  90.30  &  90.00  &  41.86  &  33.47  &  37.20  \\
      + ATLaS-ISL & 79.25  & 72.01  & 75.46  & 96.80  & 72.88  & 83.15  & 89.83  & 90.75  & \textbf{90.29} & 42.74  & 33.81  & \textbf{37.75} \\
    \midrule
    MaPLe & 76.45  & 72.03  & 74.17  & 97.18  & 70.33  & 81.60  & 90.42  & 91.57  & 90.99  & 40.34  & 32.61  & 36.07  \\
    \midrule
      + ATLaS & 75.80  & 71.78  & 73.74  & 96.84  & 72.60  & 82.99  & 90.55  & 91.59  & 91.07  & 39.70  & 33.65  & \textbf{36.43} \\
      + ATLaS-CSL & 75.96  & 72.61  & \textbf{74.25} & 96.99  & 72.46  & 82.95  & 90.73  & 91.63  & 91.18  & 37.69  & 35.11  & 36.35  \\
     + ATLaS-ISL &  74.71  &  73.50  &  74.10  &  96.17  &  74.07  &  \textbf{83.69} &  90.63  &  91.85  &  \textbf{91.24} &  38.42  &  33.89  &  36.01  \\
    \midrule
    PromptSRC & 78.10  & 75.10  & 76.57  & 97.98  & 76.93  & 86.19  & 90.65  & 91.53  & 91.09  & 43.22  & 36.91  & 39.82  \\
    \midrule
     + ATLaS &  78.21  &  75.36  &  \textbf{76.76} &  97.91  &  77.56  &  \textbf{86.55} &  90.69  &  91.54  &  91.11  &  42.60  &  37.55  &  39.92  \\
      + ATLaS-CSL & 77.90  & 75.60  & 76.73  & 97.98  & 77.40  & 86.48  & 90.73  & 91.54  & \textbf{91.13} & 42.29  & 37.37  & 39.68  \\
      + ATLaS-ISL & 76.64  & 75.57  & 76.10  & 97.27  & 76.97  & 85.94  & 90.70  & 91.53  & 91.11  & 41.82  & 38.63  & \textbf{40.16} \\
    \bottomrule
    \end{tabular}}}
  \label{appendix_tab:base2new_2}%
\end{table*}%

\begin{table*}[htbp]
  \caption{The performance of various prompt tuning methods with or without ATLaS, ATLaS-CSL, and ATLaS-ISL on the SUN397, DTD, EuroSAT, and UCF101 datasets under the base-to-new generalization setting.}
\resizebox{\linewidth}{!}{
\setlength{\tabcolsep}{4.7mm}{
  \centering
    \begin{tabular}{L{2cm}|ccc|ccc|ccc|ccc}
    \toprule
    \multicolumn{1}{c|}{\multirow{2}{*}{Methods}} & \multicolumn{3}{c|}{SUN397} & \multicolumn{3}{c|}{DTD} & \multicolumn{3}{c|}{EuroSAT} & \multicolumn{3}{c}{UCF101} \\
\cmidrule{2-13}          & Base  & New   & H     & Base  & New   & H     & Base  & New   & H     & Base  & New   & H \\
    \midrule
    CoOp  & 80.60  & 65.89  & 72.51  & 79.44  & 41.18  & 54.24  & 92.19  & 54.74  & 68.69  & 84.69  & 56.05  & 67.46  \\
    \midrule
      + ATLaS & 81.13  & 65.87  & 72.71  & 78.93  & 44.57  & 56.97  & 91.87  & 61.27  & 73.51  & 84.50  & 57.20  & 68.22  \\
      + ATLaS-CSL & 79.90  & 72.33  & 75.93  & 79.20  & 46.43  & 58.54  & 91.77  & 63.70  & 75.20  & 83.67  & 62.57  & 71.60  \\
     + ATLaS-ISL &  78.70  &  77.33  &  \textbf{78.01} &  75.13  &  56.53  &  \textbf{64.52} &  87.73  &  65.90  &  \textbf{75.26} &  81.77  &  75.63  &  \textbf{78.58} \\
    \midrule
    CoCoOp & 79.74  & 76.86  & 78.27  & 77.01  & 56.00  & 64.85  & 87.49  & 60.04  & 71.21  & 82.33  & 73.45  & 77.64  \\
    \midrule
      + ATLaS & 79.27  & 77.33  & \textbf{78.29} & 76.20  & 59.43  & 66.78  & 85.50  & 63.33  & 72.76  & 81.03  & 74.83  & 77.81  \\
      + ATLaS-CSL & 78.70  & 77.47  & 78.08  & 73.83  & 61.07  & 66.85  & 86.03  & 64.17  & 73.51  & 80.40  & 75.80  & \textbf{78.03} \\
     + ATLaS-ISL &  78.75  &  77.63  &  78.19  &  75.04  &  60.67  &  \textbf{67.09} &  83.23  &  68.24  &  \textbf{74.99} &  80.02  &  76.04  &  77.98  \\
    \midrule
    DePT  & 82.35  & 75.67  & 78.87  & 83.45  & 57.57  & \textbf{68.14} & 88.05  & 64.43  & 74.41  & 85.47  & 73.39  & 78.97  \\
    \midrule
      + ATLaS & 82.32  & 76.69  & 79.41  & 83.33  & 54.55  & 65.94  & 88.74  & 73.19  & \textbf{80.22} & 85.99  & 74.71  & 79.95  \\
      + ATLaS-CSL & 82.14  & 77.11  & 79.55  & 82.76  & 54.99  & 66.08  & 88.97  & 72.04  & 79.61  & 85.52  & 74.49  & 79.62  \\
     + ATLaS-ISL &  82.29  &  78.08  &  \textbf{80.13} &  83.02  &  57.00  &  67.59  &  89.13  &  72.44  &  79.92  &  85.76  &  77.75  &  \textbf{81.56} \\
    \midrule
    IVLP  & 81.75  & 75.67  & 78.59  & 82.83  & 53.10  & 64.71  & 95.37  & 67.51  & 79.06  & 86.54  & 75.73  & 80.77  \\
    \midrule
      + ATLaS & 81.93  & 76.16  & 78.94  & 83.76  & 52.94  & \textbf{64.88} & 95.66  & 67.98  & 79.48  & 87.33  & 72.78  & 79.39  \\
     + ATLaS-CSL &  82.15  &  76.94  &  79.46  &  82.95  &  53.22  &  64.84  &  95.11  &  72.37  &  \textbf{82.20} &  86.00  &  77.09  &  81.30  \\
      + ATLaS-ISL & 82.04  & 78.20  & \textbf{80.07} & 81.60  & 53.10  & 64.33  & 96.17  & 65.14  & 77.67  & 86.47  & 77.02  & \textbf{81.47} \\
    \midrule
    MaPLe & 81.37  & 77.29  & 79.28  & 81.68  & 56.36  & 66.70  & 95.52  & 68.87  & 80.03  & 85.25  & 75.48  & 80.07  \\
    \midrule
      + ATLaS & 81.62  & 77.50  & 79.51  & 82.14  & 58.09  & 68.05  & 95.26  & 72.47  & 82.32  & 85.14  & 76.87  & 80.79  \\
      + ATLaS-CSL & 81.68  & 77.94  & 79.77  & 79.78  & 59.46  & 68.14  & 95.06  & 72.92  & 82.53  & 85.95  & 77.05  & 81.26  \\
     + ATLaS-ISL &  81.02  &  79.23  &  \textbf{80.12} &  81.48  &  58.74  &  \textbf{68.27} &  95.00  &  73.91  &  \textbf{83.14} &  84.99  &  78.55  &  \textbf{81.64} \\
    \midrule
    PromptSRC & 82.53  & 78.45  & 80.44  & 83.10  & 62.16  & 71.12  & 92.80  & 72.07  & 81.13  & 87.28  & 78.07  & 82.42  \\
    \midrule
     + ATLaS &  82.66  &  78.95  &  80.76  &  82.87  &  62.28  &  71.11  &  92.78  &  74.51  &  \textbf{82.65} &  86.83  &  79.00  &  \textbf{82.73} \\
      + ATLaS-CSL & 82.69  & 78.83  & 80.71  & 83.49  & 62.56  & \textbf{71.53} & 92.71  & 74.22  & 82.44  & 87.03  & 78.83  & \textbf{82.73} \\
      + ATLaS-ISL & 82.13  & 79.60  & \textbf{80.85} & 83.22  & 62.60  & 71.45  & 93.40  & 73.87  & 82.49  & 87.51  & 78.29  & 82.64  \\
    \bottomrule
    \end{tabular}}}
  \label{appendix_tab:base2new_3}%
\end{table*}%
